\documentclass[sigconf]{acmart}

\usepackage{amsmath, amssymb}
\usepackage{graphicx}
\usepackage{algorithm}
\usepackage{algorithmic}
\usepackage{url}
\usepackage{float}
\usepackage{booktabs}
\usepackage{siunitx}
\usepackage{multirow}
\usepackage{tabularx}
\usepackage{makecell}
\usepackage{subcaption}

\AtBeginDocument{%
  }

\copyrightyear{2026}
\acmYear{2026}
\setcopyright{cc}
\setcctype{by}
\acmConference[WWW '26]{Proceedings of the ACM Web Conference 2026}{April 13--17, 2026}{Dubai, United Arab Emirates}
\acmBooktitle{Proceedings of the ACM Web Conference 2026 (WWW '26), April 13--17, 2026, Dubai, United Arab Emirates}
\acmPrice{}
\acmDOI{10.1145/3774904.3792432}
\acmISBN{979-8-4007-2307-0/2026/04}

\begin{document}

\title{From Representation to Clusters: A Contrastive Learning Approach for Attributed Hypergraph Clustering}

\author{Li Ni}
\affiliation{%
 \department{School of Computer Science and Technology}
  \institution{Anhui University}
  \city{Hefei}
  \country{China}
}
\email{nili@ahu.edu.cn}

\author{Shuaikang Zeng}
\affiliation{%
 \department{School of Computer Science and Technology}
  \institution{Anhui University}
  \city{Hefei}
  \country{China}
}
\email{e24301218@stu.ahu.edu.cn}

\author{Lin Mu}
\authornote{Corresponding author.}
\affiliation{%
 \department{School of Computer Science and Technology}
  \institution{Anhui University}
  \city{Hefei}
  \country{China}
}
\email{mulin@ahu.edu.cn}

\author{Longlong Lin}
\affiliation{%
 \department{School of Computer Science and Technology}
  \institution{Southwest University}
  \city{Chongqing}
  \country{China}}
\email{longlonglin@swu.edu.cn}

\renewcommand{\shortauthors}{Li Ni, Shuaikang Zeng, Lin Mu, and Longlong Lin}

\begin{abstract}
Contrastive learning has demonstrated strong performance in attributed hypergraph clustering. Typically, existing methods based on contrastive learning first learn node embeddings and then apply clustering algorithms, such as k-means, to these embeddings to obtain the clustering results.
However, these methods lack direct clustering supervision, risking the inclusion of clustering-irrelevant information in the learned graph.
To this end, we propose a Contrastive learning approach for Attributed Hypergraph Clustering (CAHC), an end-to-end method that simultaneously learns node embeddings and obtains clustering results. 
CAHC consists of two main steps:  representation learning and cluster assignment learning. 
The former employs a novel contrastive learning approach that incorporates both node-level and hyperedge-level objectives to generate node embeddings.
The latter joint embedding and clustering optimization to refine these embeddings by clustering-oriented guidance and obtains clustering results simultaneously.
Extensive experimental results demonstrate that CAHC outperforms baselines on eight datasets.

\end{abstract}



\maketitle

\section{Introduction}
 Hypergraphs are an effective tool for modeling high-order relationships among entities,  as each hyperedge can connect several nodes \cite{BrettoHypergraph2013,ZhangAccelerating2025}.
This ability makes hypergraphs particularly useful in modeling complex relationships in various real-world applications, such as recommender systems \cite{XiaSelf-Supervised2021}, computer vision \cite{YuAdaptive2012}, and neuroscience  \cite{XiaoMulti-Hypergraph2020}. 
These high-order relationships cannot be effectively represented by ordinary pairwise graphs, where each edge connects only two nodes  \cite{berge1984hypergraphs,BrettoHypergraph2013,ZhangAccelerating2025,ZHAO2026112463}. 
Hypergraphs often contain cluster structure \cite{KeCommunity2020}, and identifying such structures has widespread applications in fields \cite{AgarwalBeyond2005}, such as parallel computing  
\cite{KabiljoSocial2017} and circuit design  \cite{KarypisMultilevel1999}.
As a result, researchers have developed various hypergraph clustering methods \cite{WangASimple2025,XiaoHypergraph2025,LiHypergraph2024}.

\begin{figure}[t!]
  \centering 
  \begin{subfigure}{\columnwidth} 
    \centering
    \includegraphics[width=0.95\textwidth]{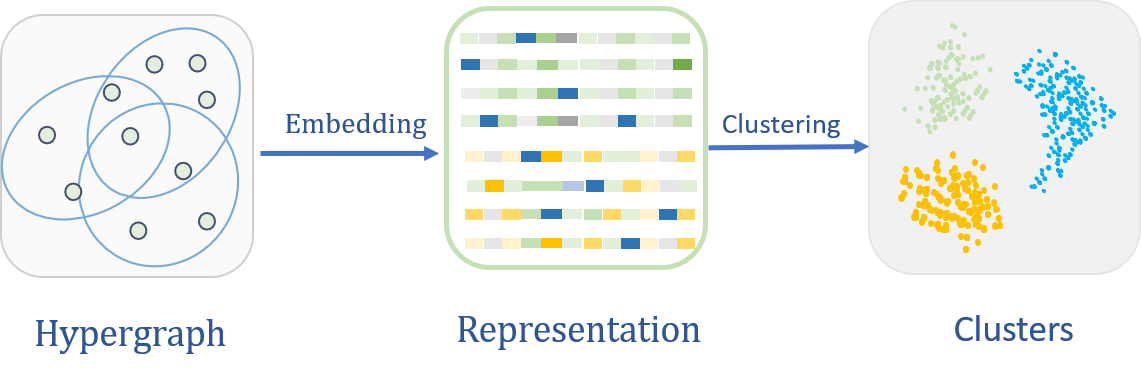} 
    \caption{Existing contrastive hypergraph clustering methods.}
    \label{fig:sub_a_label}
  \end{subfigure}
  \begin{subfigure}{\columnwidth} 
    \centering
    \includegraphics[width=0.95\textwidth]{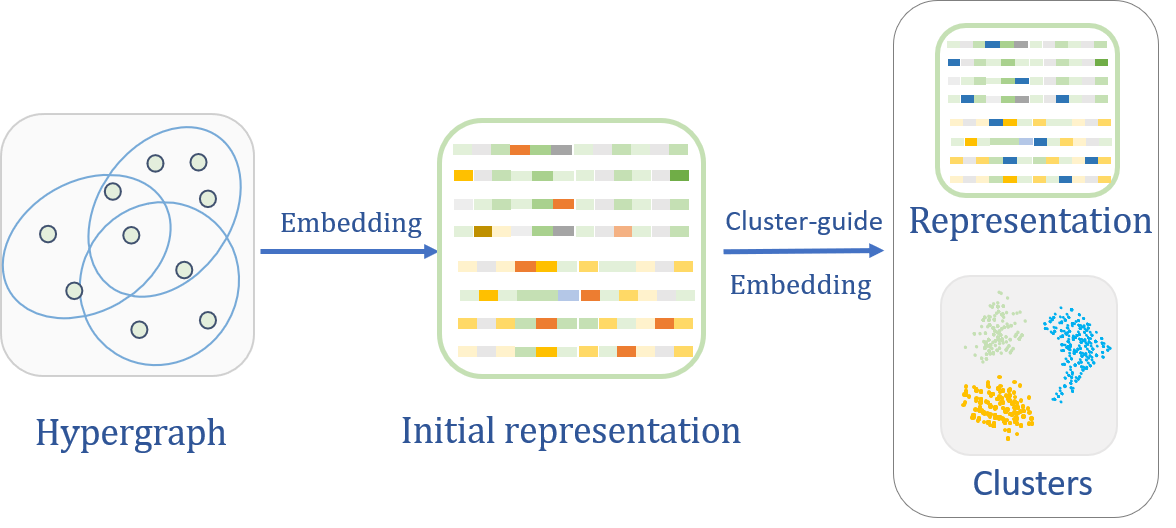}
    \caption{Proposed CAHC.}
    \label{fig:sub_b_label}
  \end{subfigure}

  \caption{Comparison of contrastive learning-based hypergraph clustering methods and the proposed CAHC.}
  \label{fig:main_framework_label}
\end{figure}

Existing hypergraph clustering methods can be broadly categorized into traditional and deep learning-based approaches  \cite{WangASimple2025}. 
Traditional methods include spectral clustering-based methods \cite{ScholkopfLearning2007,YangGraphLSHC2021}, 
fuzzy memberships-based methods  \cite{XiaoHypergraph2025},  and stochastic block models \cite{ZhangSparse2022,RuggeriFramework2024,ContiscianiInference2022}. 
These methods often struggle to capture complex and high-order structural information effectively \cite{LiHypergraph2024}.
In contrast, deep learning-based approaches have gained much attention for their ability to capture such high-order information \cite{WangASimple2025}.
Several recent self-supervised frameworks, including SE-HSSL \cite{LiHypergraph2024}, HyperGCL \cite{WeiAugmentations2022}, and TriCL \cite{LeeIm2023} have improved representation capabilities.
These methods generally use contrastive learning to learn representations, which are then partitioned using clustering algorithms, such as k-means, to form clusters, as shown in Figure  \ref{fig:main_framework_label}(\subref{fig:sub_a_label}). 

Although contrastive learning-based hypergraph clustering methods have achieved promising results, these methods typically learn representations by comparing two augmented views, lacking explicit clustering guidance. As a result, they may incorporate irrelevant or non-cluster-related features into the learned graph, ultimately resulting in low-quality clusters. 
To address this, we propose a contrastive learning approach for attributed hypergraph clustering, called CAHC. As shown in Figure \ref{fig:main_framework_label}(\subref{fig:sub_b_label}), CAHC comprises two steps: the representation learning and the cluster assignment learning. The former aims to learn high-quality node embeddings based on the hypergraph structure and attributes. The latter jointly optimizes both the embeddings and clustering, achieving a synergistic optimization of both the node embeddings and clustering results. Notably, in CAHC, traditional clustering algorithms such as k-means are no longer required.
The main contributions of this work are summarized as  follows:
 \begin{itemize}
 \item We propose an attributed hypergraph clustering method based on contrastive learning, termed CAHC. To the best of our knowledge, it is the first end-to-end model for attributed hypergraph clustering.
 CAHC employs contrastive learning for node embeddings, followed by a joint optimization of embedding and clustering objectives to yield the clustering results.

 \item   
We propose a clustering loss function that measures the closeness between soft assignments and hard assignments. It shares node embeddings with representation learning to effectively implement cluster-guidance embedding and cluster assignment simultaneously. In addition, we design a novel hyperedge-level objective to capture the structural information of hypergraphs.

 \item Extensive experiments on eight real-world datasets demonstrate that CAHC  outperforms baselines. 
 Furthermore, ablation studies confirm that each component of CAHC plays a crucial role.
 \end{itemize}

\textbf{Reproduction.} The source code of this paper can be found at https://github.com/nilics/CAHC.
\section{Proposed method}
\subsection{Problem statement}

\textbf{Data model. } We denote a hypergraph \( \mathcal{H} = (V, E, X) \), where \( V = \{ v_1, v_2, \dots, v_N \} \) represents the set of nodes, with \( N \) being the total number of nodes. The set \( E = \{ e_1, e_2, \dots, e_M \} \) represents the collection of hyperedges, where each hyperedge \( e_j \subseteq V \) is a non-empty subset of nodes and $M$ is the number of hyperedges.
Each node \( v_i \in V \) is associated with a feature vector \( \mathbf{x}_i \in \mathbb{R}^d \), which is stacked into the node feature matrix \( X \in \mathbb{R}^{N \times d} \), where each row \( X[i, :] \) corresponds to the feature vector of node \( v_i \).
The hyperedges are also represented by the incident matrix \( H \in \{0, 1\}^{N \times M} \), where \( H_{ij} = 1 \) if node \( v_i \) is part of hyperedge \( e_j \), and \( 0 \) otherwise.
Based on the incident matrix \( H \), the hypergraph \( \mathcal{H} = (V, E, X) \) can also be expressed as \( \mathcal{H} = (X, H) \).

\textbf{Problem statement.} Given a hypergraph \( \mathcal{H} = (X, H) \), the task aims to simultaneously encode the nodes in an unsupervised manner using a hypergraph neural network, while also partitioning the hypergraph into $k$ clusters \( C = \{C_1, C_2, \dots, C_k\} \) without any supervision. 

\subsection{ Overview}
\begin{figure*}[!t] 
  \centering
  \includegraphics[width=0.85\textwidth,height=4.2cm]{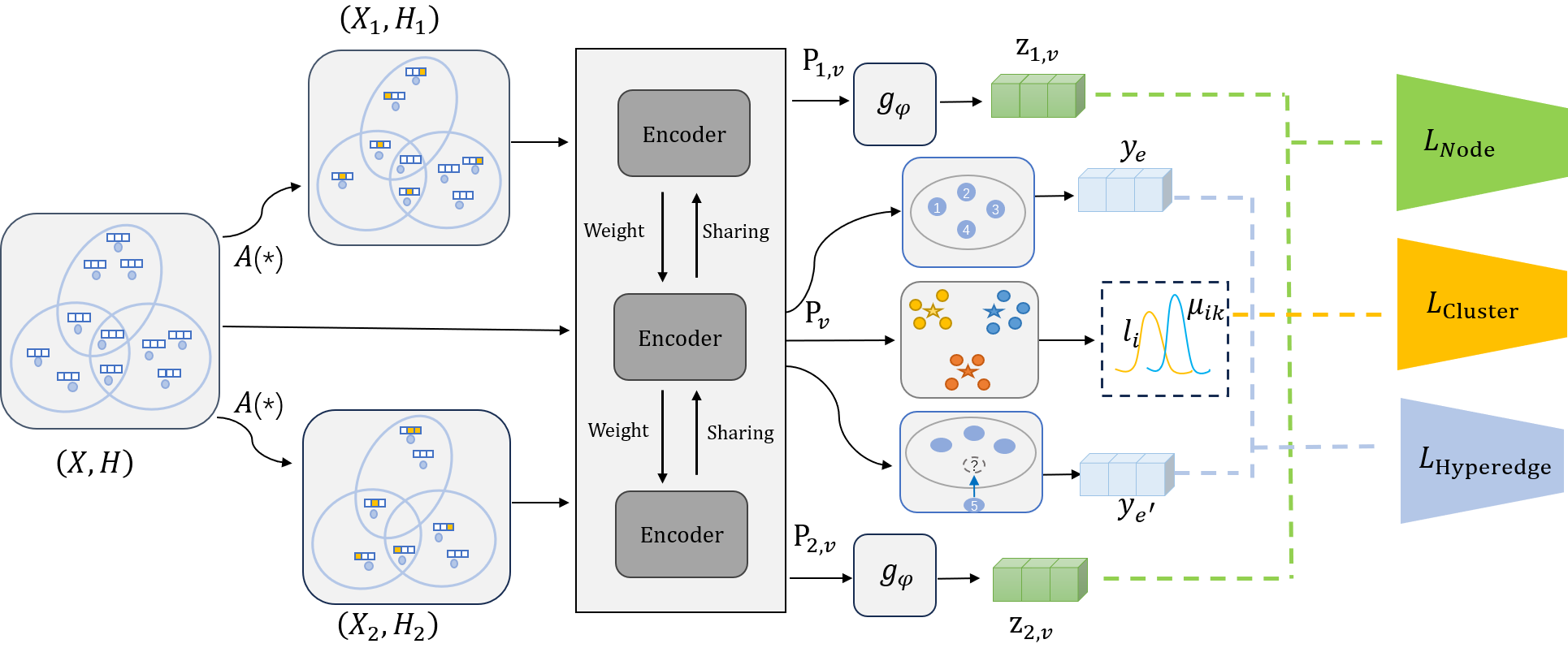} 
  \caption{ Overview of CAHC. CAHC consists of two core steps: hypergraph representation learning and cluster assignment learning.
  Specifically, hypergraph representation learning optimizes  $\mathcal{L}_{\text{hyper}} + \mathcal{L}_{\text{node}} $ to learn node embeddings from the inherent structural and attribute information of the data.
  Cluster assignment learning  optimizes  $\mathcal{L}_{\text{hyper}} $ + $\mathcal{L}_{\text{node}}$ +  $\mathcal{L}_{\text{clu}}$ to  
  achieve representations  and clustering results. }
  \Description{A flowchart illustrating the two stages of the CAHC model. Stage 1 shows data augmentation and encoding. Stage 2 shows cluster assignment learning.}
  \label{fig:framework}
\end{figure*}

In this section, we propose a contrastive learning
approach for attributed hypergraph clustering, called CAHC.
The process of  CAHC is shown in Figure \ref{fig:framework}.
CAHC comprises two core steps:  representation learning, which aims to learn node embeddings from the inherent structural and attribute information; and cluster assignment learning, which optimizes these embeddings to align with the final cluster structure, thereby achieving synergistic optimization of representation learning and clustering tasks. 

\subsection{Representation learning }
CAHC adopts a contrastive learning paradigm that learns representations by maximizing the agreement between different views.
As depicted in Figure \ref{fig:framework},  it comprises three main components: data augmentation, hypergraph encoding, and objective optimization.
 First, data augmentation is applied to the hypergraph $\mathcal{H}=({X}, {H})$ to obtain two correlated views $\mathcal{H}_1=({X}^{(1)}, {H} ^{(1)})$ and $\mathcal{H}_2=({X}^{(2)}, {H} ^{(2)})$.
Next, a shared hypergraph neural network encoder $f_{\theta}$ generates node representations $Z$ for these augmented views. 
Finally, the encoder is optimized using two learning objectives: a hyperedge-level contrastive objective and a node-level contrastive objective.  

\textbf{Data augmentation.} We consider two hypergraph augmentation strategies: node feature masking and membership relation masking. In node feature masking, we construct a binary masking matrix ${M}_f \in \{0, 1\}^{N \times d}$ with the same shape as the node feature matrix ${X}$\cite{YouGraph2020, ThakoorLarge-Scale2022}.  
Each element \(M_{f;ij}\) is independently sampled from a Bernoulli distribution: \(M_{f;ij} \sim \mathcal{B}(1 - p_{f})\), where \(p_f\) represents the probability of masking a feature element\cite{LeeIm2023}.
The augmented feature matrix ${X}'$ is computed as follows:
\begin{equation}{X}' = {X} \odot {M}_f,
\end{equation} 
where $\odot$ denotes element-wise multiplication.  This operation preserves feature values where \(M_{f;ij} = 1\), and masks (sets to zero) features where \(M_{f;ij} = 0\).

Membership relation masking is designed to perturb higher-order relationships by randomly removing or adding nodes in hyperedges. Similar to the node feature masking strategy, we construct a binary masking matrix ${M}_m \in \{0, 1\}^{N \times M}$  with the shape of the incident matrix ${H}$. 
Each element ${M}_{m, ij}$ is independently sampled from a Bernoulli distribution $\mathcal{B}(1 - p_m)$, where \(p_m\) denotes the probability of masking a node-hyperedge association\cite{LiHypergraph2024}.
The enhanced hypergraph topology is the modified incident matrix ${H}'$:
\begin{equation}
{H}' = {H} \odot {M}_m,
\end{equation}
where $\odot$ denotes element-wise multiplication. 
 This operation preserves the elements where $\mathbf{M}_{m, ij} = 1$, while masking those where $\mathbf{M}_{m, ij} = 0$. The hyperparameter $p_m$ controls the intensity of this topological perturbation.

\textbf{Hypergraph encoder.} 
The hypergraph encoder is based on a hypergraph neural network (HGNN). 
To capture the varying importance of nodes within a hyperedge, we incorporate a multi-head attention mechanism into the HGNN.
Formally, the $l$th layer of the hypergraph neural network can be expressed as:
\begin{align}
P^{(l)} &= \sigma \left(
    \mathrm{Agg}_{E \rightarrow V}(P^{(l-1)}, Q^{(l-1)}, H)\Theta_V^{(l)}
    + P^{(l-1)}\Phi_V^{(l)}
\right), \\
Q^{(l)} &= \sigma \left(
    \mathrm{Agg}_{V \rightarrow E}(P^{(l-1)}, Q^{(l-1)}, H)\Theta_E^{(l)}
    + Q^{(l-1)}\Phi_E^{(l)}
\right),
\end{align}
where
 $\sigma$ is the nonlinear activation function ReLU,
 $\Theta_V$, $\Phi_V$, $\Theta_E$, and $\Phi_E$ are learnable linear transformation matrices.
$P^{(l)}$ and ${Q}^{(l)}$ represent node and hyperedge resentations, respectively.
${P}^{(0)} $ is initialized with node features $X$, and ${Q}^{(0)}$ is initialized by aggregating the node features $X$ of its associated nodes (e.g., via mean pooling). 
The functions \( \mathrm{Agg}_{V \rightarrow E} \) and \( \mathrm{Agg}_{E \rightarrow V} \) are aggregation operations, where \( \mathrm{Agg}_{E \rightarrow V} \) refers to hyperedge-to-node aggregation and \( \mathrm{Agg}_{V \rightarrow E} \) corresponds to node-to-hyperedge aggregation. 

Aggregation functions $\mathrm{Agg}_{V \rightarrow E}$ and $\mathrm{Agg}_{E \rightarrow V}$ are based on multi-head attention mechanisms to assign different weights to distinct node and hyperedge representations adaptively. 
This mechanism addresses a key limitation of HGNNs, which aggregate information via simple averaging and thus fail to model the varying importance of nodes within a hyperedge. 
Taking the hyperedge-to-node aggregation \( \text{Agg}_{E \rightarrow V} \) as an example, it is computed as follows:
\begin{align}
\text{Agg}_{E \rightarrow V}({P}, {Q}, {H}) &= \underset{h=1}{\overset{N_h}{||}} \left( \alpha_{EV}^{(h)} {Q}{W}_Q^{(h)} \right){W}_O,
\end{align} 
where $||$ denotes the concatenation operation, $N_h$ is the number of attention heads,  ${W}_Q^{(h)}$, and ${W}_O$ are learnable weight matrices. The attention coefficient matrix $\alpha_{EV}^{(h)}$ is computed as follows to learn the attention of node $v_i$ to its associated hyperedge $e_j$:
\begin{align}
\alpha_{EV, ij}^{(h)} &= \mathrm{softmax}_j(s_{ij}^{(h)}) = \frac{\exp(s_{ij}^{(h)})}{\sum_{j' \in \mathcal{N}_V(i)} \exp(s_{ij'}^{(h)})}, \\
s_{ij}^{(h)} &= \mathrm{LeakyReLU} \left( ({a}^{(h)})^\top [{p}_i {W}_P^{(h)} || {q}_j {W}_Q^{(h)}] \right),
\end{align}
where $\mathcal{N}_V(i)$ denotes the set of hyperedges associated with node $v_i$. The term $s_{ij}^{(h)}$ represents the unnormalized attention score of node $v_i$ towards hyperedge $e_j$ under the $h$-th attention head, which is parameterized by a learnable attention vector $\mathbf{a}^{(h)}$. The aggregation from nodes to hyperedges, denoted as $\text{Agg}_{V\rightarrow E}$, is implemented symmetrically.

\textbf{Contrastive loss.} 
Based on the aforementioned data augmentation strategy and shared hypergraph encoder, we optimize the encoder by minimizing the following two loss functions: hyperedge-level loss ${L}_{\text{hyper}} $ and node-level loss ${L}_{\text{node}}$.

The hyperedge-level loss aims to enable the model to distinguish which hyperedges are real or generated negative hyperedges, as follows:
\begin{equation} 
\label{eq:hyper_loss}
    \mathcal{L}_{\text{hyper}} = - \frac{1}{|E|} \sum_{e \in E}\log(\sigma(S({y}_e))) - \frac{1}{|E_{\text{neg}}|} \sum_{e \in E_{\text{neg}}} \log(1 - \sigma(S({y}_{e} ))),
\end{equation}
where $\sigma(\cdot)$ denotes the Sigmoid function, $S(\cdot)$ represents a shared scoring function  (implemented as a multilayer perceptron) that computes a scalar score for all hyperedge representations, \( {y}_e \) is the representation of hyperedge \( e \), obtained by applying mean pooling to the node embeddings (generated by the encoder) that belong to \( e \), $E$ represent hyperedges in the original hypergraph, and \( E_{\text{neg}} \) represents the set of generated negative hyperedges.
These negative hyperedges are constructed by randomly replacing nodes in each original hyperedge 
\( e \in E \) to form invalid ones.
The hyperedge-level loss adheres to the principle of contrastive learning by comprising two opposing objectives: a positive term that maximizes the similarity for nodes connected by real hyperedges in \( E \), and a negative term that minimizes the similarity for nodes connected by artificially constructed negative hyperedges in \( E_{\text{neg}} \).

Node-level loss aims to make the representation of the same node in two augmented views similar, while ensuring that these representations are distinguishable from those of other nodes.
For augmented  views $\mathcal{H}_1=({X}^{(1)}, {H} ^{(1)})$ and $\mathcal{H}_2=({X}^{(2)}, {H} ^{(2)})$, the symmetric node loss function $\mathcal{L}_{\text{node}}$ is computed as follows:
\begin{equation}
  \label{eq:node_loss}
    \mathcal{L}_{\text{node}} = \frac{1}{2N} \sum_{i=1}^{N} \left[ \ell({z}_i^{(1)} , {z}_i^{(2)})+ \ell({z}_i^{(2)}, {z}_i^{(1)}) \right],
\end{equation}
where  $\ell({z}_i^{(1)}, {z}_i^{(2)})$ is computed as:
\begin{equation}
  \label{eq:node_loss_single}
    \ell({z}_{i}^{(1)}, {z}_{i}^{(2)}) = -\log \frac{\exp(\text{sim}({z}_{i}^{(1)}, {z}_{i}^{(2)})/\tau_n)}{\sum_{k=1}^{N} \exp(\text{sim}({z}_{i}^{(1)}, {z}_{k}^{(2)})/\tau_n)},
\end{equation}
where $\text{sim}(\cdot,\cdot)$ denotes cosine similarity, and $\tau_n$ is a temperature hyperparameter, ${z}_{i}^{(1)}$ and ${z}_{i}^{(2)}$ represent the final node representation vectors obtained for the same node $v_i$ in two augmented views, respectively, which through a shared encoder $f_{\theta}$ and a nonlinear projection head $g_{\varphi}$.
Here, for any node representation ${z}_{i}^{(1)}$ from the first view, its corresponding representation ${z}_{i}^{(2)}$ in the second view forms a positive sample pair. Meanwhile, ${z}_{i}^{(1)}$ forms negative sample pairs with all other node representations ${z}_{j}^{(2)}$ ($j \neq i$) from the second view\cite{LeeIm2023}.

The hyperedge-level loss calculated by Eq. (\ref{eq:hyper_loss}) and node-level loss calculated by Eq. (\ref{eq:node_loss}) are complementary. The hyperedge loss forces the model to understand the complex, higher-order interaction patterns between nodes, focusing on the structural within the hypergraph. 
On the other hand, the node loss ensures that the representations of different nodes are sufficiently distinct, emphasizing the individual discriminability of nodes. Together, these losses are integrated into the final contrastive loss,   calculated as:
\begin{equation} \label{eq:contra_loss}
    \mathcal{L}_{\text{rep}} = \mathcal{L}_{\text{hyper}} + \mathcal{L}_{\text{node}}.
\end{equation}
 CAHC jointly optimizes Eq.  (\ref{eq:contra_loss}) during the representation learning phase, enabling the learned node embeddings to simultaneously preserve information at both the hyperedge and node levels.

\subsection{Cluster assignment learning }
Embedding methods without clustering supervision may learn non-cluster-related features, resulting in poor clustering results.
To provide clustering-oriented guidance, CAHC performs joint optimization of the embedding and clustering, thereby obtaining the representation and the final cluster assignments simultaneously.

CAHC assumes that there are \(K\) clusters, denoted by $c = \{c_1, c_2, $ $\dots,$ $ c_K\}$. 
It computes the membership degree \(\mu_{ik}\) for each node \(i\) with respect to each cluster center \(c_k\), forming a soft assignment matrix. It also assigns each node to its nearest cluster center, generating high-confidence pseudo-labels \(l_i\) (i.e., hard assignments). 
The loss function is designed to minimize the discrepancy between the  soft assignment   and hard assignments, as follows:
\begin{equation}
\label{eq:clus_loss}
    \mathcal{L}_{\text{clus}} = -\frac{1}{N}\sum_{i=1}^{N}\sum_{k=1}^{K} \mathbb{I}(l_i=k)\log \mu_{ik},
\end{equation}
where $\mathbb{I}(\cdot)$ denotes the indicator function and  soft assignment of node $i$ to cluster  ${c}_k$ is computed as follows:
\begin{equation}
\label{eq:soft_assignment}
    \mu_{ik} = \frac{\exp(\text{sim}({z}_i, {c}_k)/\tau_c)}{\sum_{k'=1}^{K}\exp(\text{sim}({z}_i, {c}_{k'})/\tau_c)},
\end{equation}
where $\tau_c$ is a temperature hyperparameter,   ${z}_i$  represents  node embedding of node $i$,
and cluster centers $c = \{c_1, c_2, $ $\dots,$ $ c_K\}$ are obtained by k-means \cite{HartiganA1979}. 
The  pseudo-label $l_i$ is computed as:
\begin{equation}
\label{eq:pseudo_label}
    l_i = \arg\max_k \mu_{ik}.
\end{equation}

\begin{algorithm}[!t]
\caption{CAHC }
\label{alg:CAHC}
\begin{algorithmic}[1]
\REQUIRE Node features $\mathbf{X}$, hypergraph incidence matrix $\mathbf{H}$, number of clusters $K$, representation learning epochs $T_{1}$, cluster assignment learning epochs $T_{2}$, loss weights $\alpha, \beta, \gamma$.
\ENSURE Final node cluster assignments $\mathbf{Y}$.

\STATE \textbf{Stage 1: Representation Learning}
\FOR{$epoch = 1$ \TO $T_{1}$}
    \STATE  Generate two augmented views $\mathcal{H}_1$ and $\mathcal{H}_2$.
    \STATE Construct negative hyperedges $E_{\text{neg}}$ 
     \STATE Compute loss $\mathcal{L}_{\text{rep}}$ with  Eq.~\eqref{eq:contra_loss}.
    \STATE Update  the encoder $f_{\theta}$ and projection head $g_{\varphi}$ 
\ENDFOR
\STATE  Obtain node representation  $\mathbf{Z}_{\text{init}}$
\STATE \textbf{Stage 2: Cluster assignment learning}
\STATE  Cluster centroids $\mathbf{c}$ $\leftarrow$ k-means ($\mathbf{Z}_{\text{init}}$)
\FOR{$epoch = 1$ \TO $T_{2}$}
 \STATE Compute loss $\mathcal{L}_{\text{rep}}$ with  Eq.~\eqref{eq:contra_loss}.
    \STATE Compute the soft assignments $\mathbf{Q}$ with Eq.~\eqref{eq:soft_assignment}
     \STATE Compute  pseudo-labels $\hat{\mathbf{Y}}$ with Eq.~\eqref{eq:pseudo_label}
    \STATE Compute  clustering  loss $\mathcal{L}_{\text{clus}}$ with Eq.~\eqref{eq:clus_loss}
    \STATE Update all model parameters  by $\mathcal{L}$ with Eq.~\eqref{eq:loss}
\ENDFOR

\STATE Obtain  clustering results with $\mathbf{Y}$  from Eq. (\ref{eq:pseudo_label})
\RETURN $\mathbf{Y}$
\end{algorithmic}
\end{algorithm}

The overall loss function $\mathcal{L}$ combines the clustering loss $\mathcal{L}_{\text{clus}}$ in Eq. (\ref{eq:clus_loss}) with the representation learning loss $\mathcal{L}_{\text{rep}}$ in Eq. (\ref{eq:contra_loss}):
\begin{equation}
\label{eq:loss}
    \mathcal{L} = \mathcal{L}_{\text{clus}}+ \mathcal{L}_{\text{rep}}.
\end{equation}

Algorithm \ref{alg:CAHC} summarizes the execution steps of CAHC. 
CAHC generates two augmented views to compute the node loss $\mathcal{L}_{\text{node}}$ according to Eq.~\eqref{eq:node_loss}. It constructs negative hyperedges $E_{\text{neg}}$ to compute the hypergraph loss $\mathcal{L}_{\text{hyper}}$ in Eq.~\eqref{eq:hyper_loss}  and then computes the contrastive loss.
During the learning phase, CAHC optimizes Eq.~\eqref{eq:contra_loss} to update the encoder $f_{\theta}$ and the projection head $g_{\varphi}$, learning representations. 
Based on the representations $Z_{\text{init}}$, the algorithm applies k-means to obtain the cluster centroids.
Using $Z_{\text{init}}$ and the cluster centroids, CAHC proceeds to optimize Eq.~\eqref{eq:loss}, performing joint embedding and clustering optimization to obtain both the learned representations and the clustering results. The clustering results are derived using the pseudo-labels $l_i$, which are computed according to Eq.~\eqref{eq:pseudo_label}.

\section{Experiment}
In this section, we first introduce the experimental settings, including the datasets, baselines, and evaluation metrics. Subsequently, we compare CAHC with the baselines, analyze the impact of clustering guidance, conduct ablation studies, and examine parameter sensitivity.

\subsection{Experiment settings}

\textbf{Datasets}. We evaluate the performance of the algorithms on eight public datasets: Cora-C \cite{SenCollective2008}, Citeseer \cite{SenCollective2008}, Pubmed \cite{SenCollective2008}, Cora-A \cite{RossiThe2015}, DBLP \cite{RossiThe2015}, NTU2012 \cite{ChenASimple2020}, 20NewsW100 \cite{DuaUCI2017}, and Mushroom \cite{DuaUCI2017}. 
Statistics information about these datasets is shown in Table \ref{tab:datasets_statistics_transposed}. For details regarding the descriptions of each dataset, refer to Appendix~\ref {sec:appendix_dataset_details}.

\begin{table}[!t]
  \centering
  \caption{Statistics of the datasets.}
  \label{tab:datasets_statistics_transposed}
  \small 
  \begin{tabular}{lrrrrr}
    \toprule
    \textbf{Dataset} & \textbf{\# Nodes} & \textbf{\# Hyperedges} & \textbf{\# Features} & \textbf{\# Classes} \\
    \midrule
    Cora-C & 1,434 & 1,579 & 1,433 & 7 \\
    Citeseer & 1,458 & 1,079 & 3,703 & 6 \\
    Pubmed & 3,840 & 7,963 & 500 & 3 \\
    Cora-A & 2,388 & 1,072 & 1,433 & 7 \\
    DBLP   & 41,302 & 22,363 & 1425 & 6 \\
    20NewsW100 & 16,242 & 100 & 100 & 4 \\
    Mushroom & 8,124 & 298 & 22 & 2 \\
    NTU2012 & 2,012 & 2,012 & 100 & 67 \\
    \bottomrule
  \end{tabular}
\end{table}


\textbf{Baselines}. We compare CAHC with six baselines: three classical embedding methods (Node2vec \cite{GroverNode2vec2016}, DGI  \cite{VelickovicDeep2019} and Hyper2vec  \cite{HuangHyper2vec2019}), a graph learning method ( RAGC  \cite{Zhao2025RAGC}), and two hypergraph self-supervised learning methods (TriCL  \cite{LeeIm2023} and SE-HSSL  \cite{LiHypergraph2024}).
Referring to  \cite{LeeIm2023,LeeVilLain2024}, for self-supervised methods designed for graphs (DGI and RAGC), we first convert hypergraphs to graphs via clique expansion and apply these methods to the converted graphs. 
For embedding-based baseline methods, the learned embeddings are fed into the k-means algorithm \cite{HartiganA1979} for clustering. Refer to  Appendix~\ref{sec:appendix_implementation_details} for the detailed settings of CAHC and baselines.

\textbf{Metrics}.
We employ four metrics to assess clustering performance: Accuracy (ACC) \cite{MiFast2024}, Normalized Mutual Information (NMI) \cite{ZhouMultiple2020}, Adjusted Rand Index (ARI) \cite{WangLate2023}, and Macro-F1 Score \cite{WangLate2023}.
For ACC and F1 scores, we utilize the Hungarian algorithm for label alignment \cite{KuhnThe1955,XieUnsupervised2016}. 
For all metrics, higher values indicate better clustering performance.

\subsection{Experiment result}


Table 2 demonstrates the clustering performance of all algorithms on eight public datasets.
Overall, CAHC outperforms baselines on most datasets.
It is observed that methods designed for hypergraphs (e.g., TriCL, SE-HSSL, and CAHC) outperform those designed for graphs (Node2Vec, DGI). Since the latter requires first converting hypergraphs to ordinary graphs via clique expansion, this conversion leads to information loss.
Notably, CAHC outperforms TriCL and SE-HSSL on the Citeseer, Pubmed, and Mushroom datasets. Specifically, on the Pubmed dataset, CAHC achieves relative improvements of 10.3\% and 17.1\% in NMI and ARI, respectively, compared to TriCL and SE-HSSL. This is primarily because TriCL and SE-HSSL rely on a two-step process: first, learning embeddings, followed by clustering, which lacks effective clustering guidance. In contrast, our algorithm optimizes both representation learning and clustering in a unified, end-to-end manner.

RAGC gets the best performance on the Mushroom dataset, which contains only two clusters.
Due to the small number of clusters, RAGC’s contrastive sample adaptive-differential awareness module generates highly accurate pseudo-labels. This reliability allows the module to effectively discriminate and weight difficult samples, leading to well-optimized cluster boundaries. In contrast, on datasets with more clusters, such as Cora-C and Citeseer, the initial pseudo-labels are less accurate.
On the 20NewsW100 dataset, the performance of CAHC is inferior to that of SE-HSSL. The negative hyperedge generation strategy employed by CAHC is not well-suited for hyperedges of large size, where each hyperedge connects, on average, more than 160 nodes in 20NewsW100. Specifically, the strategy of replacing only a single node to generate negative hyperedges fails to produce effective negative samples. This is because the generated 'negative samples' are nearly identical to the original samples, thus diminishing the discriminative power of the hyperedge-level loss.

\begin{table*}[t!]
  \caption{Comparison of clustering results (\%) on datasets. All results are reported as mean ± std over 5 independent runs. The best results are in \textbf{bold}. ‘OOM’ denotes the out-of-memory failure, while ‘OOT’ signifies no results are obtained within 24 hours.}
  \label{tab:main_results_acm_style}
  \centering
  \normalsize 
  \renewcommand{\arraystretch}{1.2} 
  \newcommand{\score}[2]{{\large #1}{\small $\pm$#2}}
  \newcolumntype{C}{>{\centering\arraybackslash}X}
  \setlength{\tabcolsep}{4pt} 
  \begin{tabularx}{\textwidth}{ ll *{7}{C} }
    \toprule
    \textbf{Dataset} & \textbf{Metric} & \textbf{Node2vec} & \textbf{DGI} & \textbf{RAGC} & \textbf{Hyper2vec} & \textbf{TriCL} & \textbf{SE-HSSL} & \textbf{CAHC} \\
    \midrule
    \multirow{4}{*}{\textbf{Cora-C}} 
    & ACC & \score{47.4}{5.4} & \score{68.4}{3.4} & \score{68.6}{1.6} & \score{59.7}{5.1} & \score{70.0}{1.1} & \textbf{\score{72.0}{3.5}} & \score{71.4}{1.7} \\
    & F1  & \score{44.7}{7.1} & \score{62.0}{3.0} & \score{64.8}{1.6} & \score{58.5}{0.4} & \score{65.1}{1.8} & \score{67.0}{4.3} & \textbf{\score{69.1}{2.8}} \\
    & NMI & \score{32.5}{3.9} & \score{54.0}{1.0} & \score{50.9}{1.7} & \score{40.6}{1.1} & \score{55.2}{0.8} & \textbf{\score{56.5}{2.4}} & \score{56.4}{0.9} \\
    & ARI & \score{20.2}{4.2} & \score{47.1}{2.0} & \score{43.5}{2.2} & \score{33.6}{0.9} & \score{49.2}{0.2} & \score{49.3}{3.5} & \textbf{\score{50.8}{1.6}} \\
    \midrule
    \multirow{4}{*}{\textbf{Citeseer}} 
    & ACC & \score{36.2}{1.6} & \score{63.8}{4.3} & \score{69.4}{1.4} & \score{42.7}{1.3} & \score{67.6}{1.5} & \score{68.2}{1.7} & \textbf{\score{69.8}{0.8}} \\
    & F1  & \score{29.2}{3.1} & \score{55.7}{3.4} & \score{61.9}{1.9} & \score{41.1}{1.2} & \score{62.1}{2.0} & \score{62.5}{1.5} & \textbf{\score{65.1}{0.6}} \\
    & NMI & \score{20.0}{1.5} & \score{40.0}{2.1} & \score{43.5}{0.6} & \score{21.0}{1.0} & \score{44.1}{0.4} & \score{43.8}{2.1} & \textbf{\score{45.9}{1.2}} \\
    & ARI & \score{5.7}{0.6} & \score{38.6}{2.8} & \score{44.3}{1.7} & \score{16.9}{1.1} & \score{45.1}{0.3} & \score{43.3}{2.0} & \textbf{\score{46.2}{1.2}} \\
    \midrule
    \multirow{4}{*}{\textbf{Pubmed}} 
    & ACC & \score{53.8}{0.5} & \score{63.9}{8.0} & \score{67.0}{2.5} & \score{69.4}{5.4} & \score{62.4}{0.6} & \score{65.8}{2.3} & \textbf{\score{69.4}{0.2}} \\
    & F1  & \score{54.6}{0.5} & \score{61.8}{10.6} & \score{65.4}{2.2} & \score{67.7}{5.0} & \score{61.3}{0.6} & \score{65.1}{2.4} & \textbf{\score{69.0}{0.2}} \\
    & NMI & \score{21.6}{0.4} & \score{27.0}{6.4} & \score{27.1}{2.4} & \score{29.6}{3.3} & \score{29.7}{0.6} & \score{31.2}{1.1} & \textbf{\score{34.4}{0.4}} \\
    & ARI & \score{12.1}{0.5} & \score{24.9}{8.0} & \score{26.6}{3.1} & \textbf{\score{33.5}{6.1}} & \score{24.7}{0.8} & \score{27.5}{2.2} & \score{32.2}{0.3} \\
    \midrule
    \multirow{4}{*}{\textbf{Cora-A}}
    & ACC & \score{34.0}{2.0} & \score{60.9}{5.0} & \score{59.7}{1.1} & \score{22.7}{0.2} & \score{70.2}{1.5} & \score{61.0}{3.9} & \textbf{\score{72.6}{0.4}} \\
    & F1  & \score{24.4}{2.2} & \score{53.5}{6.4} & \score{54.9}{1.8} & \score{17.8}{0.1} & \score{69.1}{1.2} & \score{52.3}{2.5} & \textbf{\score{70.6}{0.4}} \\
    & NMI & \score{12.5}{1.2} & \score{44.1}{1.6} & \score{42.2}{0.3} & \score{3.0}{0.5} & \score{50.2}{1.2} & \score{47.6}{1.9} & \textbf{\score{53.1}{0.7}} \\
    & ARI & \score{3.1}{1.2} & \score{37.0}{3.6} & \score{34.5}{0.9} & \score{0.6}{0.2} & \score{45.8}{2.5} & \score{38.8}{3.6} & \textbf{\score{50.1}{0.7}} \\
    \midrule
    \multirow{4}{*}{\textbf{DBLP}}
    & ACC & \score{39.91}{3.3} & \multirow{4}{*}{OOM} & \multirow{4}{*}{OOM} & \multirow{4}{*}{OOM} & \score{65.9}{4.1} & \multirow{4}{*}{OOT} & \textbf{\score{72.3}{8.0}} \\
    & F1  & \score{38.05}{3.8} & & & & \score{63.7}{5.3} & & \textbf{\score{70.4}{8.6}} \\
    & NMI & \score{24.20}{1.9} & & & & \score{62.4}{3.7} & & \textbf{\score{63.9}{2.8}} \\
    & ARI & \score{9.63}{1.3} & & & & \score{50.4}{5.2} & & \textbf{\score{58.2}{5.9}} \\
    \midrule
    \multirow{4}{*}{\textbf{20NewsW100}}
    & ACC & \score{44.6}{1.2} & \multirow{4}{*}{OOM} & \multirow{4}{*}{OOM} & \multirow{4}{*}{OOM} & \score{61.5}{3.7} & \textbf{\score{70.2}{0.7}} & \score{64.6}{3.1} \\
    & F1  & \score{41.8}{0.4} & & & & \score{54.9}{2.7} & \textbf{\score{65.0}{1.0}} & \score{63.2}{2.8} \\
    & NMI & \score{11.1}{0.7} & & & & \score{36.0}{3.8} & \score{39.1}{1.4} & \textbf{\score{39.5}{1.1}} \\
    & ARI & \score{12.1}{1.0} & & & & \score{29.4}{9.4} & \textbf{\score{40.4}{0.2}} & \score{35.6}{2.5} \\
    \midrule
    \multirow{4}{*}{\textbf{Mushroom}}
    & ACC & \score{69.9}{6.5} & \score{70.7}{2.5} & \textbf{\score{87.5}{0.8}} & \multirow{4}{*}{OOM} & \score{51.3}{1.4} & \score{78.3}{6.7} & \score{80.5}{2.4} \\
    & F1  & \score{69.3}{6.8} & \score{67.8}{3.1} & \textbf{\score{87.4}{0.8}} & & \score{38.8}{2.4} & \score{77.4}{7.6} & \score{80.5}{2.5} \\
    & NMI & \score{15.0}{6.8} & \score{20.9}{4.8} & \textbf{\score{46.6}{2.8}} & & \score{3.4}{0.6} & \score{28.7}{10.5} & \score{30.3}{4.2} \\
    & ARI & \score{17.6}{8.2} & \score{17.3}{3.8} & \textbf{\score{56.3}{2.5}} & & \score{0.5}{1.0} & \score{34.2}{14.4} & \score{37.5}{5.9} \\
    \midrule
    \multirow{4}{*}{\textbf{NTU2012}}
    & ACC & \score{56.7}{2.2} & \score{65.0}{2.1} & \score{66.1}{0.3} & \score{58.8}{0.6} & \score{70.5}{0.5} & \score{70.5}{0.6} & \textbf{\score{70.8}{1.1}} \\
    & F1  & \score{51.4}{1.7} & \score{59.8}{2.0} & \score{57.4}{1.1} & \score{54.0}{0.6} & \score{63.8}{0.8} & \textbf{\score{64.0}{0.7}} & \score{63.7}{1.1} \\
    & NMI & \score{77.3}{0.4} & \score{80.0}{0.6} & \score{79.9}{0.6} & \score{78.1}{0.1} & \score{83.1}{0.2} & \score{83.2}{0.3} & \textbf{\score{83.8}{0.2}} \\
    & ARI & \score{50.1}{2.5} & \score{59.4}{3.2} & \score{63.1}{1.2} & \score{51.8}{0.6} & \score{66.8}{0.5} & \score{66.9}{1.5} & \textbf{\score{67.0}{2.0}} \\
    \bottomrule
  \end{tabularx}
\end{table*}

\subsection{Impact of clustering guidance}
To investigate the effect of the clustering guidance, we design two variants of CAHC: (1) ``CAHC + k-means". 
This variant generates embeddings from CAHC and applies k-means to them to obtain the final results; and (2) ``w/o Clus".  This variant first obtains embeddings from the representation learning step and then applies k-means to these embeddings to obtain the results.
We evaluate CAHC + k-means, w/o Clus, and the original CAHC on the Pubmed, Cora-A, and DBLP datasets.

\begin{figure*}[t!]
  \centering
  \begin{subfigure}[b]{0.33\textwidth}
    \centering
    \includegraphics[height=3.5cm, keepaspectratio]{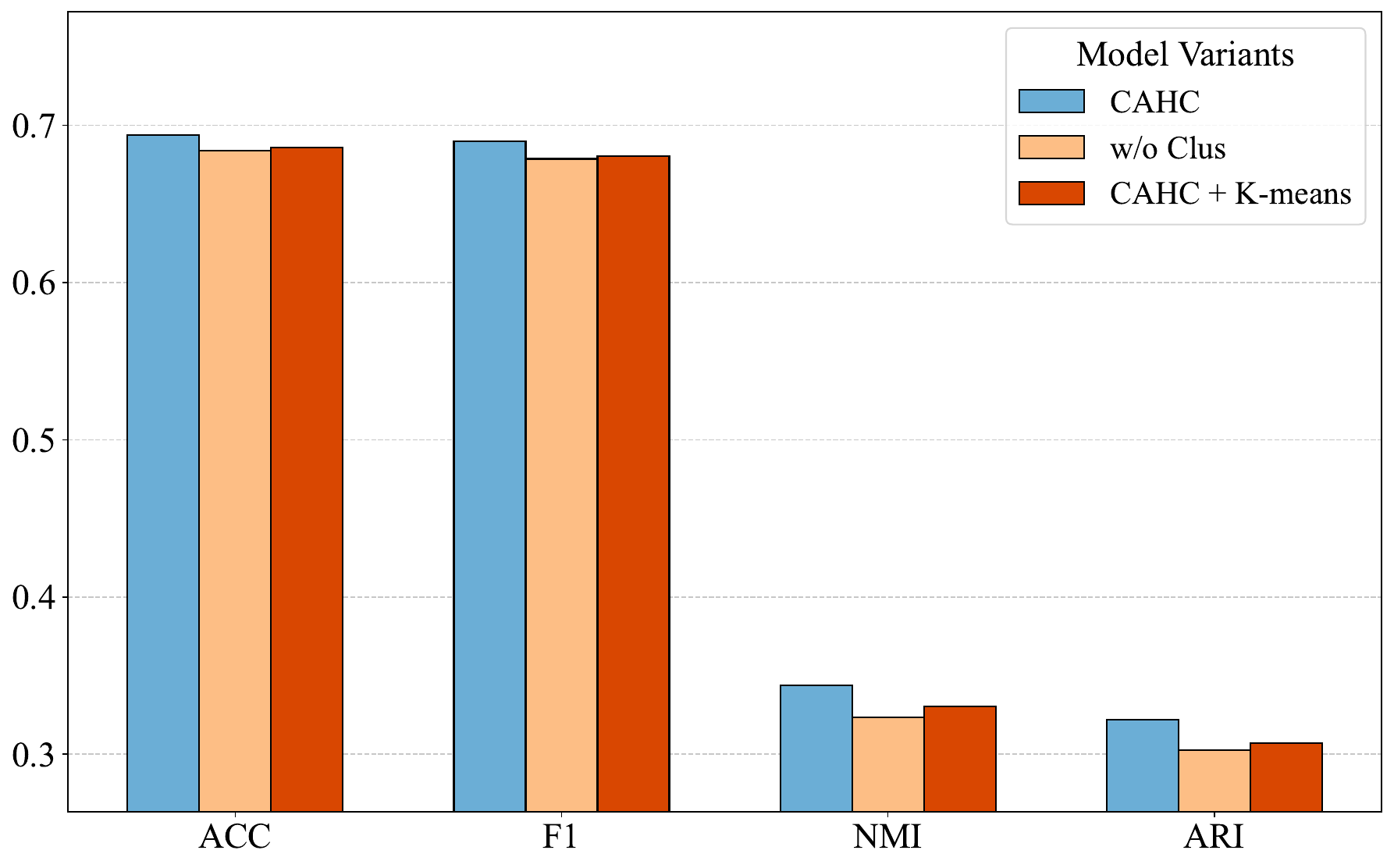}
    \caption{Pubmed}
    \label{fig:compare_pubmed}
  \end{subfigure}
  \hfill 
  \begin{subfigure}[b]{0.33\textwidth}
    \centering
    \includegraphics[height=3.5cm, keepaspectratio]{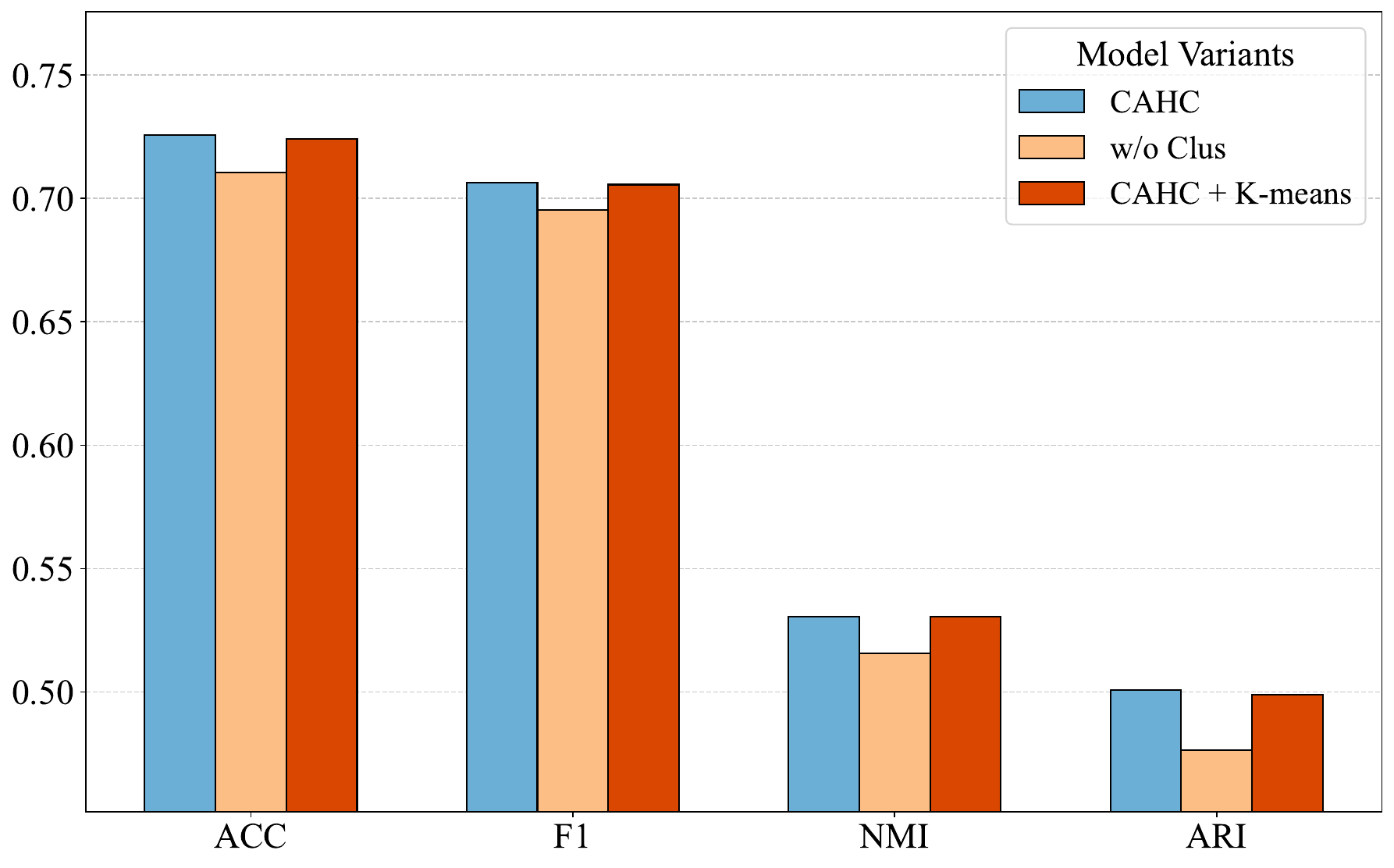}
    \caption{Cora-A}
    \label{fig:compare_cora_coauthor}
  \end{subfigure}
  \hfill 
  \begin{subfigure}[b]{0.33\textwidth}
    \centering
    \includegraphics[height=3.5cm, keepaspectratio]{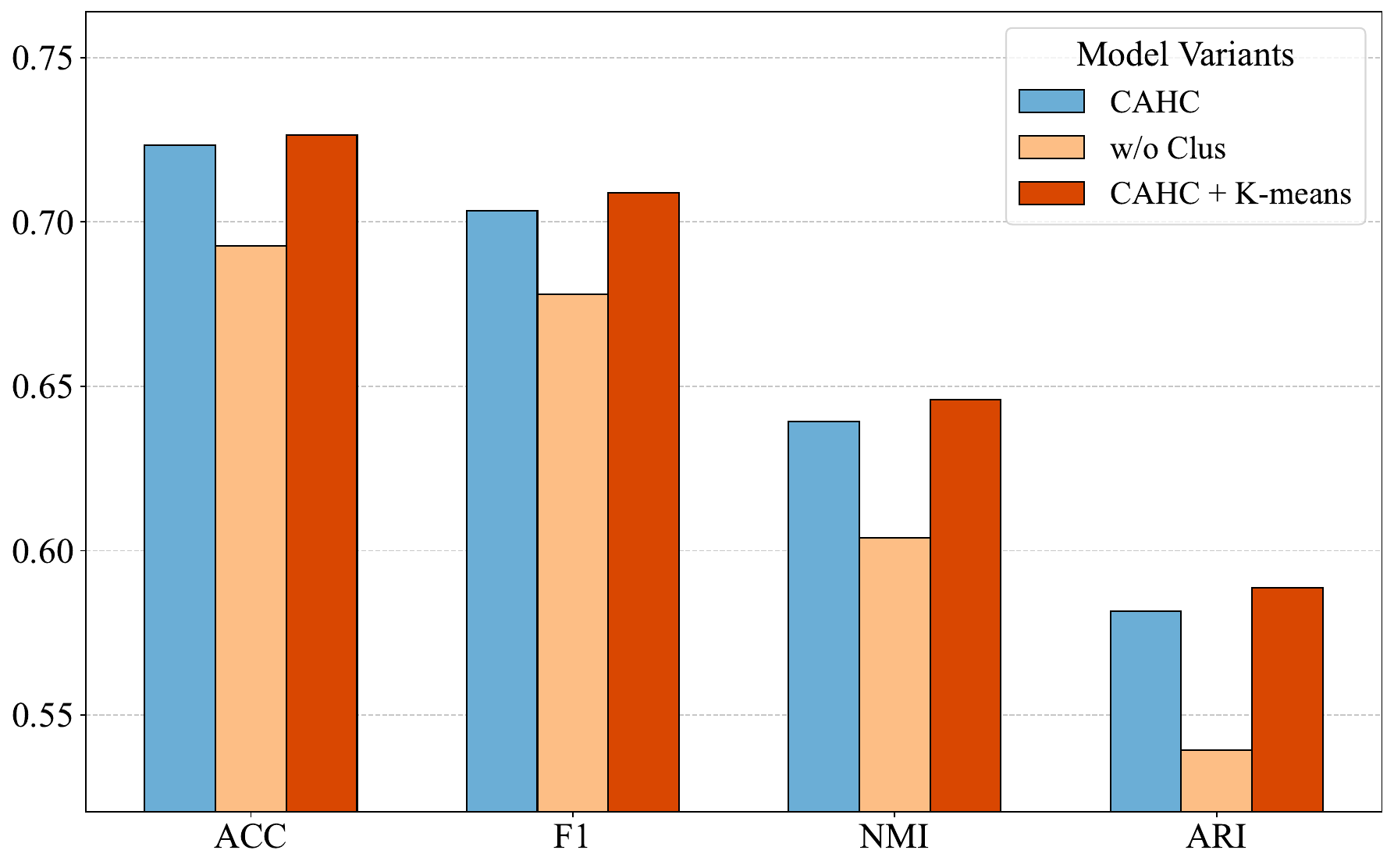}
    \caption{DBLP}
    \label{fig:comparison_dblp_coauthor}
  \end{subfigure}
  
  \caption{ Impact of clustering guidance.}
  \label{fig:key_dataset_comparisons}
\end{figure*}

Figure \ref{fig:key_dataset_comparisons} presents ACC, F1, NMI, and ARI of CAHC + k-means,  w/o Clus, and  CAHC. 
Figure~\ref{fig:key_dataset_comparisons} shows that CAHC + k-means outperforms w/o Clus, demonstrating that the clustering loss helps improve the embedding quality for the clustering task. Furthermore, Figure~\ref{fig:key_dataset_comparisons} indicates that `CAHC + k-means` and CAHC are comparable. This is because the membership score $\mu_{ik}$, calculated by Eq.~(\ref{eq:soft_assignment}), follows a similar principle as k-means, where each node is assigned to the nearest cluster center.

\subsection{Ablation study}
To validate components of  CAHC, we conducted an ablation study on eight datasets.
Specifically, five  variants of CAHC are constructed:
1)  "w/o re". This variant represents CAHC without the representation learning step; 
2) "w/o hy". This variant represents CAHC without the hyperedge-level loss in Eq. (~\ref{eq:hyper_loss});
3) "w/o no". This variant represents CAHC without the node-level  loss in  Eq.  (~\ref{eq:node_loss}); 
4)  "w/o cl". This variant represents CAHC without the clustering module, where clustering is performed using k-means; 
and 5) "w/o mu". This variant represents CAHC without the multi-head attention mechanism, using a standard HGNN \cite{FengHypergraph2019} instead.
Table~\ref{tab:ablation_single_column_final} demonstrates the results.
Overall, CAHC outperforms w/o re, "w/o hy", "w/o no", "w/o cl", and "w/o mu", indicating that the representation learning step, hyperedge-level loss, node-level loss, clustering module, and the encoder all play a crucial role. Afterward, we gradually analyze the impact of each component.

\textbf{Effectiveness of contrastive objectives.} 
The "w/o hy", which removes hyperedge-level loss, performs significantly worse than CAHC on most datasets.
For instance, on the Pubmed dataset, CAHC achieves an NMI score of 34.4, while the score of "w/o hy" is only 29.8.  This empirically validates that explicitly learning from high-order structural information is a crucial component of the model's success.
Similarly, without the node-level contrastive loss ("w/o no") also performs poorly. This is particularly evident on  Cora-A, where the NMI score of CAHC is 53.1, while that of "w/o no" is only 45.4.
This indicates that the individual discriminability of nodes is also indispensable for node representation.

\textbf{Effectiveness of the end-to-end learning paradigm.} 
"w/o cl" without clustering guidance performs worse than CAHC, indicating that such guidance is crucial for improving representations.
"w/o cl" trains the encoder on the hypergraph without clustering guidance, leading to the incorporation of non-cluster-related features in the learned representations. Most NMI/ARI of "w/o re" are 0 because "w/o re" assigns all nodes in the dataset to a single cluster, even when the number of clusters is set to be greater than one. It indicates that the representation learning step plays a crucial role.

\textbf{Effectiveness of the proposed encoder architecture.}  
"w/o mu" with a standard HGNN performs significantly worse than CAHC.
Specifically, on the Cora-A dataset, the ARI score of "w/o mu" is 41.0, while the CAHC model achieves a score of 50.1, showing a considerable performance gap. Similarly, on the Citeseer dataset, the ARI of "w/o mu" is 42.0, and the ARI of CAHC is 45.9, indicating a clear difference in performance.
The results confirm that compared with HGNN, the multi-head attention mechanism-based HGNN could help  CAHC to learn better representations.

\begin{table}[!t]
\centering
\caption{Ablation study results (\%).}
\label{tab:ablation_single_column_final}
\small
\setlength{\tabcolsep}{2pt} 
\begin{tabular*}{\columnwidth}{@{\extracolsep{\fill}} l l c c c c c c} 
\hline
\textbf{Dataset} & \textbf{Metric} & \textbf{w/o re} & \textbf{w/o hy} & \textbf{w/o no} & \textbf{w/o cl} & \textbf{w/ mu} & \textbf{CAHC} \\
\hline
\multirow{4}{*}{\textbf{Cora-C}} 
    & ACC & 28.0 & 68.8 & \bfseries 72.9 & 70.4 & 71.6 & 71.4 \\
    & F1  & 6.2 & 63.4 & \bfseries 70.5 & 68.4 & 67.1 & 69.1 \\
    & NMI & 0.0 & 51.4 & 54.4 & 55.2 & 54.5 & \bfseries 56.4 \\
    & ARI & 0.0 & 46.5 & \bfseries 51.9 & 49.2 & 49.9 & 50.8 \\
\hline
\multirow{4}{*}{\textbf{Citeseer}} 
    & ACC & 21.7 & 69.1 & 65.5 & 68.8 & 67.0 & \bfseries 69.8 \\
    & F1  & 5.9 & 64.9 & 60.6 & 64.3 & 62.4 & \bfseries 65.1 \\
    & NMI & 0.0 & 45.0 & 42.9 & 44.5 & 42.0 & \bfseries 45.9 \\
    & ARI & 0.0 & 44.7 & 42.1 & 44.7 & 42.5 & \bfseries 46.2 \\
\hline
\multirow{4}{*}{\textbf{Pubmed}} 
    & ACC & 40.7 & 61.6 & 69.2 & 68.4 & \bfseries 70.4 & 69.4 \\
    & F1  & 19.3 & 60.4 & 68.4 & 67.9 & \bfseries 69.7 & 69.0 \\
    & NMI & 0.0 & 29.8 & 33.0 & 32.4 & 34.0 & \bfseries 34.4 \\
    & ARI & 0.0 & 25.2 & 30.5 & 30.3 & \bfseries 33.1 & 32.2 \\
\hline
\multirow{4}{*}{\textbf{Cora-A}} 
    & ACC & 30.2 & 69.8 & 65.1 & 71.1 & 65.9 & \bfseries 72.6 \\
    & F1  & 6.7 & 68.4 & 59.3 & 69.5 & 63.7 & \bfseries 70.6 \\
    & NMI & 0.0 & 51.1 & 45.4 & 51.6 & 46.2 & \bfseries 53.1 \\
    & ARI & 0.0 & 46.6 & 42.3 & 47.6 & 41.0 & \bfseries 50.1 \\
\hline
\multirow{4}{*}{\textbf{DBLP}} 
    & ACC & 27.2 & 52.4 & 68.8 & 69.3 & \bfseries 72.6 & 72.3 \\
    & F1  & 7.1 & 43.2 & 64.9 & 67.8 & \bfseries 70.8 & 70.4 \\
    & NMI & 0.0 & 52.3 & 60.9 & 60.4 & 63.0 & \bfseries 63.9 \\
    & ARI & 0.0 & 28.6 & 52.6 & 53.9 & 54.9 & \bfseries 58.2 \\
\hline
\multirow{4}{*}{\textbf{20newsW100}} 
    & ACC & 33.6 & 62.9 & 58.6 & 64.1 & 62.5 & \bfseries 64.6 \\
    & F1  & 12.6 & 61.8 & 56.7 & 63.0 & 62.0 & \bfseries 63.2 \\
    & NMI & 0.0 & \bfseries 39.6 & 33.8 & 39.2 & 36.9 & 39.5 \\
    & ARI & 0.0 & 35.0 & 29.2 & 34.5 & 29.7 & \bfseries 35.6 \\
\hline
\multirow{4}{*}{\textbf{Mushroom}} 
    & ACC & 51.8 & 72.7 & 80.4 & 80.4 & 74.3 & \bfseries 80.5 \\
    & F1  & 34.1 & 70.9 & 80.4 & 80.3 & 72.2 & \bfseries 80.5 \\
    & NMI & 0.0 & 21.2 & 30.2 & 29.8 & 26.5 & \bfseries 30.3 \\
    & ARI & 0.0 & 23.2 & 37.2 & 37.2 & 25.1 & \bfseries 37.5 \\
\hline
\multirow{4}{*}{\textbf{NTU2012}} 
    & ACC & 5.1 & \bfseries 71.0 & 69.9 & 70.6 & 69.6 & 70.8 \\
    & F1  & 0.3 & \bfseries 64.8 & 63.1 & 63.7 & 62.6 & 63.7 \\
    & NMI & 0.3 & \bfseries 83.8 & 83.4 & 83.7 & 82.4 & \bfseries 83.8 \\
    & ARI & 0.0 & 66.8 & 65.5 & 66.7 & 66.2 & \bfseries 67.0 \\
\hline
\end{tabular*}
\end{table}

\subsection{Parameter sensitivity analysis}
Here, we investigate the sensitivity of hyperparameters:  feature masking rate $p_{\text{f}}$,  the membership relation masking rate $p_{\text{m}}$, and embedding dimension $D$.

\begin{figure}[t!]
  \centering
  \begin{subfigure}[b]{0.65\columnwidth}
    \centering
    \includegraphics[width=\textwidth]{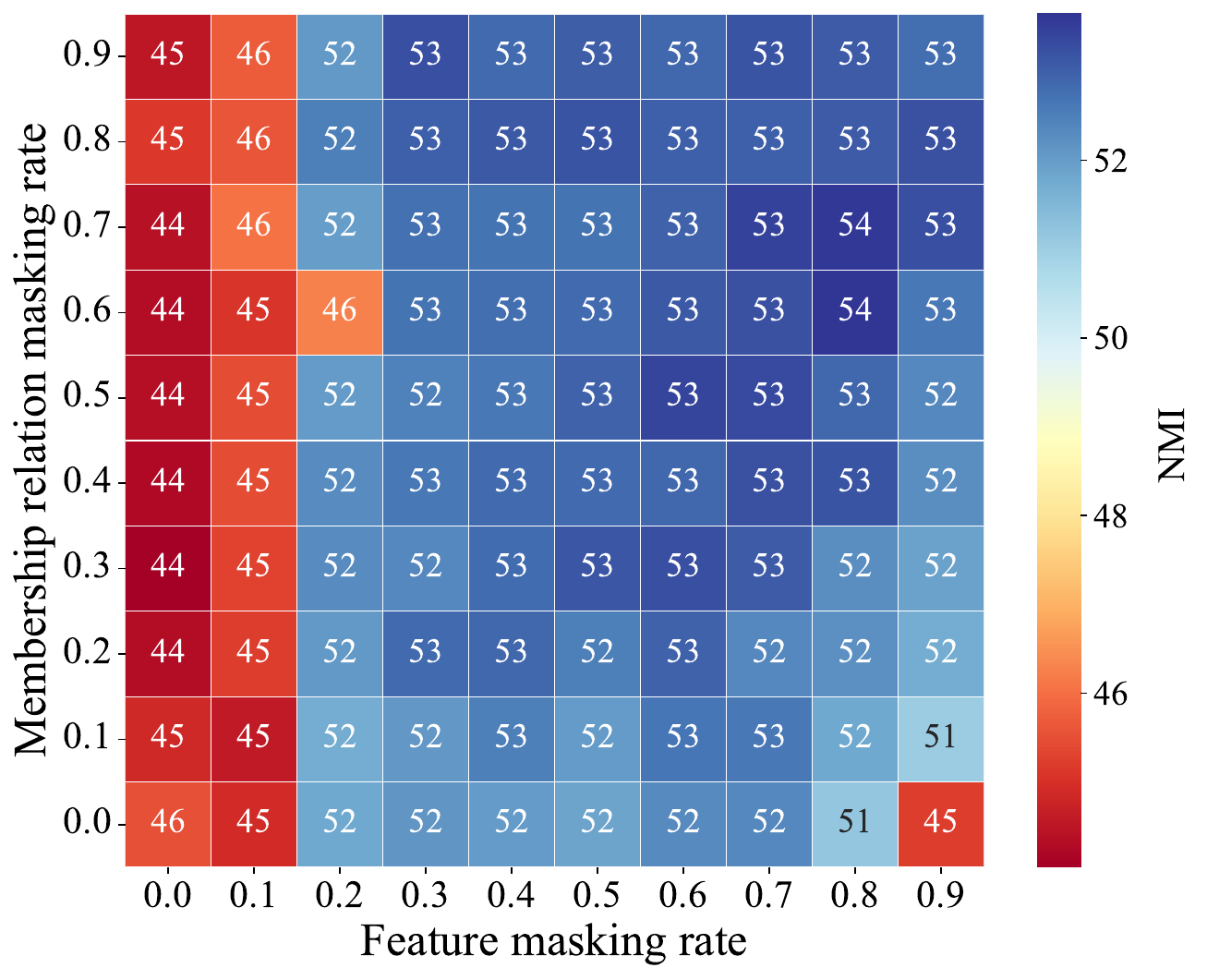} 
    \caption{Cora-A}
    \label{fig:heatmap_cora_coauthor}
  \end{subfigure}
  \begin{subfigure}[b]{0.65\columnwidth}
    \centering
    \includegraphics[width=\textwidth]{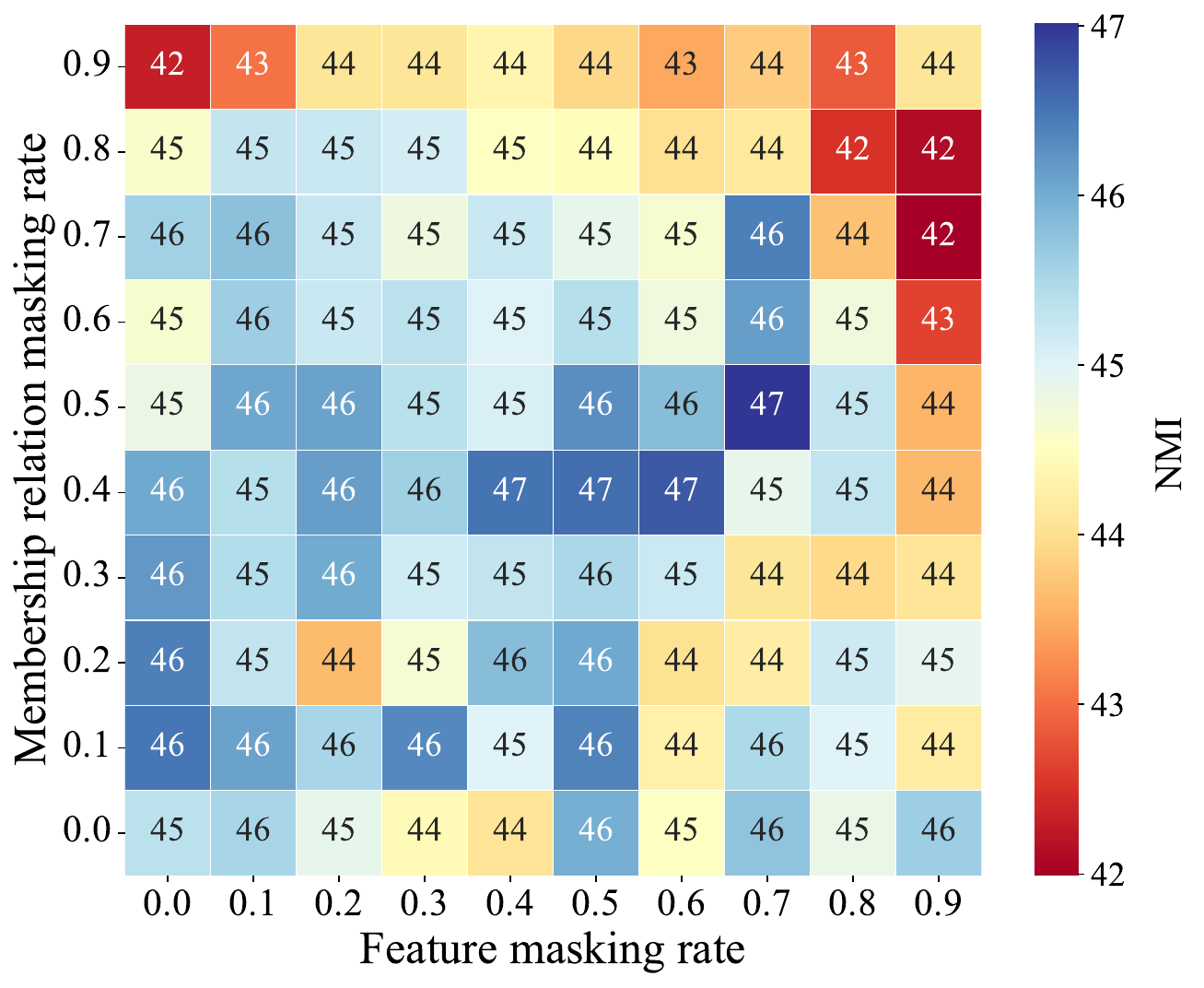}
    \caption{Citeseer}
    \label{fig:heatmap_citeseer}
  \end{subfigure}
  \caption{Sensitivity analysis of  $p_{\text{f}}$ and  $p_{\text{m}}$.
  }
  \label{fig:sensitivity_augmentation}
\end{figure}

\textbf{Effect of augmentation rates.} 
The rates \(p_{\mathrm{f}}\) and \(p_{\mathrm{m}}\) ranged from 0.0 to 0.9 with a step size of 0.1.
Figure~\ref{fig:sensitivity_augmentation} shows NMI of CAHC.
A consistent conclusion can be drawn from both Cora-A and Citeseer datasets: CAHC achieves optimal performance at moderate values of 
\(p_{\mathrm{f}}\) and \(p_{\mathrm{m}}\).
Specifically, the model achieves its highest performance when the masking rates are within an intermediate range, roughly between 0.2 and 0.7 for both parameters. If the masking rates are too small (approaching 0.0), two overly similar views are generated. This provides an insufficient learning signal for the contrastive objective to develop a robust and discriminative encoder. Conversely, if the rates are too large (approaching 0.9), the underlying feature and structural information of the original hypergraph is broken, leading to a significant drop in performance.  Therefore, intermediate augmentation rates are selected in  experiments.

\begin{figure}[t!]
    \centering
    \includegraphics[width=0.85\columnwidth]{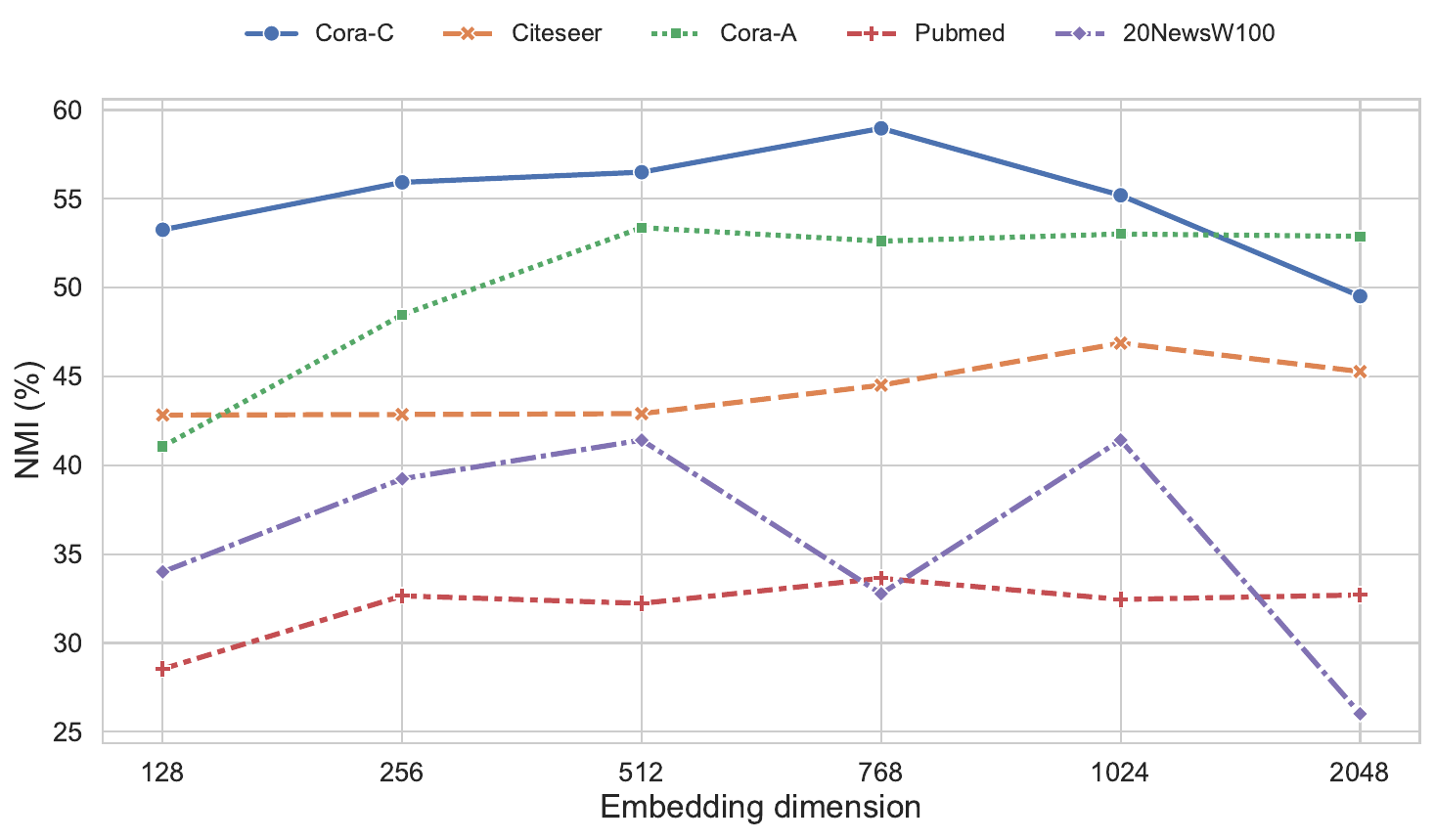}
    \caption{Sensitivity analysis of  $D$.
    }
    \label{fig:sensitivity_emb_dim}
\end{figure}

\textbf{Effect of embedding dimension.} 
The embedding dimension $D$ is chosen from the set \{128, 256, 512, 768, 1024, 2048\}.
Figure~\ref{fig:sensitivity_emb_dim} illustrates NMI of CAHC.
As $D$ increases from 128 to an intermediate value (typically 512 or 768), the performance of CAHC improves overall. This suggests that CAHC benefits from a larger representation capacity to effectively capture the complex structural and feature information. However, we notice a subsequent decline in performance as the dimension further increases on most datasets. 
Taking 20NewsW100 as an example, which has only 100 original features, the model reaches its peak performance at an embedding dimension D of 512. The performance of CAHC deteriorates significantly when the dimension is increased to 2048. 
This indicates that datasets with limited features do not necessitate high-dimensional embeddings and may, in fact, suffer from excessively large dimensions.
This phenomenon can be attributed to the embedding space becoming overly sparse, with many dimensions potentially becoming irrelevant or redundant. 
Thus, a medium-sized embedding dimension is suitable for most datasets. For datasets with high-dimensional features, a large dimension should be used.

\section{Related work}
\textbf{ Hypergraph representation learning.} Hypergraph representation learning aims to embed the nodes (or hyperedges) of a hypergraph into a low-dimensional vector space.
 \cite{AntelmiA2023,LiHySAE2025,AryaHyperSAGE2020}. 
Due to the availability of labels, many studies have focused on semi-supervised learning for hypergraph representation  \cite{WuSCHG2025,YadatiHyperGCN2019}.
  For example, Feng et al.  \cite{FengHypergraph2019}  generalize graph convolutions to the hypergraph domain, with the core idea of using the hypergraph Laplacian operator for information propagation. 
Wu et al. \cite{WuSCHG2025}  leverages spectral graph theory to capture the global latent structure within hypergraph learning.
Yan et al. \cite{YanHypergraph2024} propose an expansion method to transform a hypergraph into a standard graph and design a learning model to embed both hypervertices and hyperedges into a joint representation space.
Qu et al. \cite{QuHypergraph2026} propose a transformer-based framework that adopts a one-stage message passing paradigm.

These methods rely on labeled data as supervised signals, which may lead to high labeling costs and poor generalization. Recently, contrastive learning (CL) has gained attention for unsupervised hypergraph representation learning, as it does not require labels.
For example,
HyperGCL \cite{WeiAugmentations2022} addresses the critical issue of view construction in hypergraph contrastive learning by designing both knowledge-based manual augmentations and generative augmentation methods.
Li et al.  \cite{XiaHypergraph2022} proposed the Hypergraph Contrastive Collaborative Filtering (HCCF) framework, which integrates hypergraph structure learning with cross-view contrastive learning.
To further capture multi-granularity structural information, TriCL  \cite{LeeIm2023} employs a three-level contrastive framework involving nodes, hyperedges, and their membership relations.
These studies primarily focus on learning node representations, requiring clustering algorithms like k-means for clustering tasks. 
Different from them, our work proposes an end-to-end approach that simultaneously learns node representations and generates clustering results.

\textbf{ Attributed hypergraph clustering. } 
Attributed hypergraph clustering aims to use structure and attributes to discover communities or clusters in hypergraphs  \cite{FengQPC25,WangASimple2025,KamhouaZ0CLH21}.
Existing methods can be broadly categorized into traditional and deep learning-based methods  \cite{WangASimple2025}. 
Traditional methods include modularity-based  \cite{FengQC23}, 
fuzzy memberships-based method   \cite{XiaoHypergraph2025}, 
and  stochastic block model  \cite{ZhangSparse2022,RuggeriFramework2024,ContiscianiInference2022}. 
With the ability to capture high-order information, deep graph clustering, especially contrastive learning, has received increasing attention \cite{FengQPC25,WangASimple2025}.
For example, SE-HSSL \cite{LiHypergraph2024}, HyperGCL \cite{WeiAugmentations2022}, and TriCL \cite{LeeIm2023}  use contrastive learning to learn representations, which are then partitioned using clustering algorithms, such as k-means, to form clusters.
Additionally, Feng et al.  \cite{FengQPC25} regard a multi-hop modularity as the objective function to obtain representation and use Louvain for clustering.
Wang et al.  \cite{WangASimple2025} propose a hypergraph clustering network that utilizes hypergraph smoothing preprocessing instead of hypergraph convolution to avoid high computational complexity.  Subsequently, k-means is employed to partition low-dimensional representations into several clusters.
The above studies require clustering algorithms like k-means or Louvain to detect clusters.
Unlike these studies, our work proposes an end-to-end clustering method that simultaneously obtains node representations and clustering results.
\section{Conclusion}
In this paper, we propose an unsupervised hypergraph clustering algorithm, CAHC.
It consists of two main steps: hypergraph representation learning and cluster assignment learning. 
Hypergraph representation learning leveraging both node-level and hyperedge-level contrastive learning to obtain node embeddings. In addition, to capture the structural information of hypergraphs, we design a novel hyperedge-level objective that maximizes the similarity for nodes connected by real hyperedges and minimizes the similarity for nodes connected by constructed negative hyperedges. 
Cluster assignment learning refines these embeddings by aligning them with the final cluster structure,  obtaining the final clustering assignments.
Extensive experimental results demonstrate the consistent superiority of the proposed algorithm across various hypergraph datasets.

\section*{Acknowledgments}
This work was supported by the National Natural Science Foundation of China [No.62572002, and No.62272001] and Natural Science Foundation of Anhui Province of China [No.2508085MF159]. 

\bibliographystyle{ACM-Reference-Format}
\bibliography{ref}

\appendix

\section{Details of datasets}
\label{sec:appendix_dataset_details}
To evaluate the performance of the algorithms, we conduct experiments on eight datasets. These datasets include co-citation networks (Cora, Citeseer, and Pubmed)~ \cite{SenCollective2008}; co-authorship networks (Cora-Coauthor and DBLP-Coauthor)~ \cite{RossiThe2015}; a computer vision dataset (NTU2012)~ \cite{ChenASimple2020}; and two datasets from the UCI Machine Learning Repository (20NewsW100 and Mushroom)~ \cite{DuaUCI2017}. Statistics information about these datasets is shown in Table \ref{tab:datasets_statistics_transposed}. 

For the co-citation network, nodes represent academic papers. A hyperedge represents a set of papers that are all cited together by a common publication.
For the co-authorship network, nodes also represent papers. A hyperedge represents all papers that are written by the same author \cite{LeeIm2023}. 
For co-citation and co-authorship datasets, node features are bag-of-words representations derived from the paper abstracts, and node labels correspond to their research fields. The preprocessed hypergraphs for Cora, Citeseer,  Pubmed, Cora-Coauthor, and DBLP-Coauthor datasets are publicly available with the official implementation of HyperGCN~ \cite{YadatiHyperGCN2019, LeeIm2023}. For the visual dataset NTU2012, the hypergraph construction follows the setting described in~ \cite{FengHypergraph2019}, with node features extracted by a pre-trained convolutional neural network. For 20NewsW100 and Mushroom, node features are TF-IDF representations and categorical descriptions, respectively. 
In experiments, we delete isolated nodes in datasets. Since our work focuses on unsupervised node clustering, the entire graph structure and all node features are utilized during training, and the ground-truth labels are used exclusively for post-hoc evaluation.

\section{Implementation details}
\label{sec:appendix_implementation_details}
All experiments are performed on a system equipped with an NVIDIA GeForce RTX 4090 GPU (48GB memory), 64GB of RAM, and an Intel Core i9-13900K CPU. CAHC  is implemented using PyTorch 2.4.1 and PyTorch Geometric 2.6.1.

\subsection{Hyperparameters}
CAHC is trained using the Adam optimizer and has several parameters.
Several hyperparameters were fixed across all experiments, introduced as follows. The number of encoder layers was set to 1. Within the enhanced encoder, the number of attention heads was fixed at 4, and the dimension of each attention head was set to 128. In the appendix~\ref{sec:appendix_encoder_sensitivity}, we have added an independent sensitivity analysis regarding the number and dimension of attention heads. 
The contrastive temperature $\tau$ was set to 0.5.

Some parameters take different values on different datasets, such as the embedding dimension ($D$), learning rate (LR), cluster assignment learning learning rate (Ass.LR), weight decay, augmentation rates ($p_{\text{f}}$, $p_{\text{m}}$), and the number of epochs for representation learning(Pre.E) and cluster assignment learning(Ass.E), respectively.
The values of these parameters for each dataset are provided in Table~\ref{tab:my_hyperparameters}.

\begin{table*}[t] 
  \caption{Key hyperparameter settings for our model (CAHC). Fixed parameters are detailed in the text.}
  \label{tab:my_hyperparameters}
  \centering
  \normalsize
  \renewcommand{\arraystretch}{1.2}
  \setlength{\tabcolsep}{13.5pt} 
  \begin{tabular}{l c c c c c c c c} 
    \toprule
    \textbf{Dataset} & \textbf{$p_f$} & \textbf{$p_m$} & \textbf{D} & \textbf{LR} & \textbf{Ass. LR} & \textbf{WD} & \textbf{Pre. E} & \textbf{Ass. E} \\
    \midrule
    Cora-C & 0.4 & 0.4 & 512 & 5e-4 & 5e-4 & 5e-4 & 180 & 120 \\
    Citeseer & 0.4 & 0.4 & 2048 & 1e-4 & 1e-5  & 5e-5 & 200 & 80 \\
    Pubmed & 0.1 & 0.4 & 1024 & 5e-5 & 5e-5 & 5e-5 & 150 & 60 \\
    Cora-A & 0.3 & 0.2 & 768 & 1e-4 & 1e-5 & 1e-5 & 100 & 140 \\
    DBLP & 0.4 & 0.2 & 256 & 8e-4 & 8e-4& 5e-5 & 260 & 100 \\
    20NewsW100 & 0.1 & 0.2 & 512 & 1e-3 & 1e-4 & 5e-4 & 80 & 30 \\
    Mushroom & 0.4 & 0.2 & 128 & 1e-4 & 1e-4 & 5e-4 & 120 & 80 \\
    NTU2012 & 0.4 & 0.2 & 512 & 1e-5 & 1e-5& 5e-5 & 60 & 40 \\
    \bottomrule
  \end{tabular}
\end{table*}

  


\subsection{Baseline details}

To evaluate the performance of CAHC, we compare it with  seven baselines,  introduced as follows:

    1) Node2vec \cite{GroverNode2vec2016}. Node2vec is a graph representation learning framework that generates node sequences through a biased random walk and learns node embeddings in conjunction with the skip-gram model.
    
    2) DGI \cite{VelickovicDeep2019}. DGI is an unsupervised GCN framework that learns node embeddings by maximizing the mutual information between local node representations and the global graph summary.
    
   3) RAGC \cite{Zhao2025RAGC}. RAGC is a robust attributed graph clustering algorithm. It primarily comprises a hybrid collaborative augmentation module and a contrastive sample adaptive-differential awareness module.

   4) Hyper2vec \cite{HuangHyper2vec2019}.  Hyper2vec is a representation learning algorithm that extends the random walk paradigm to hypergraphs. It designs a biased second-order random walk strategy to sample node sequences to capture the high-order structural information, and combines it with the skip-gram framework to learn the final node embeddings. 
    
    5) TriCL \cite{LeeIm2023}.   TriCL is a three-level contrastive framework that captures multi-granular structural information from nodes, hyperedges, and their membership interactions.   k-means is employed on the node embeddings to generate clusters.
    
    6) SE-HSSL \cite{LiHypergraph2024}.   SE-HSSL incorporates sampling-free CCA-based objectives at both node and group levels, alongside a novel hierarchical membership contrastive objective.
     k-mean is employed on the node embeddings to generate clusters.

Referring to  \cite{LeeIm2023,LeeVilLain2024}, for self-supervised methods designed for graphs (DGI and RAGC), we first convert hypergraphs to graphs via clique expansion and apply these methods to the converted graphs. 
For embedding-based baseline methods, the learned embeddings are fed into the k-means algorithm \cite{HartiganA1979} for clustering.
For all baselines, we used the publicly available code from the original paper. 
Since both CAHC and the baselines are unsupervised, no labeled validation data was available for model selection. 
Thus, for each baseline, we either adopted their default hyperparameter settings or conducted a grid search to find configurations that work well on datasets, following the guidelines from papers. The parameter settings for the baselines designed for graphs and hypergraphs are described as follows, respectively.

For the baselines designed for graphs, Node2vec, DGI, and RAGC---we first transformed the hypergraphs into standard graphs using clique expansion, where each hyperedge is converted into a clique. Their specific settings are as follows. For Node2vec, we used the default hyperparameters; specifically, we set the number of walks per node to 10, the length of each walk to 80, the window size to 5, and the learning rate to 0.05, with parameters $p$ and $q$ both set to 1. For DGI, we used the PReLU activation function and set the learning rate to 0.001, as given by default. For RAGC, the hidden dimension was set to 1500, and the numbers of local and global diffusion steps ($t_l$, $t_m$) were both set to 3. The confidence threshold $\tau$ was 0.9, the positive sample weight $\beta$ was 0.9, and the negative sample weight $\gamma$ was 1.0.

For the baselines designed for hypergraphs, we used the following settings. For Hyper2vec, the number of walks was set to 10 with a length of 20, the window size was 5, and the parameters $p$ and $q$ were both set to 1. For both TriCL and SE-HSSL, we strictly adhered to the default hyperparameter settings provided in their respective official implementations and original papers, without any modifications.

\section{Scalability}

\begin{figure}[htbp!]
    \centering
    \includegraphics[width=\columnwidth]{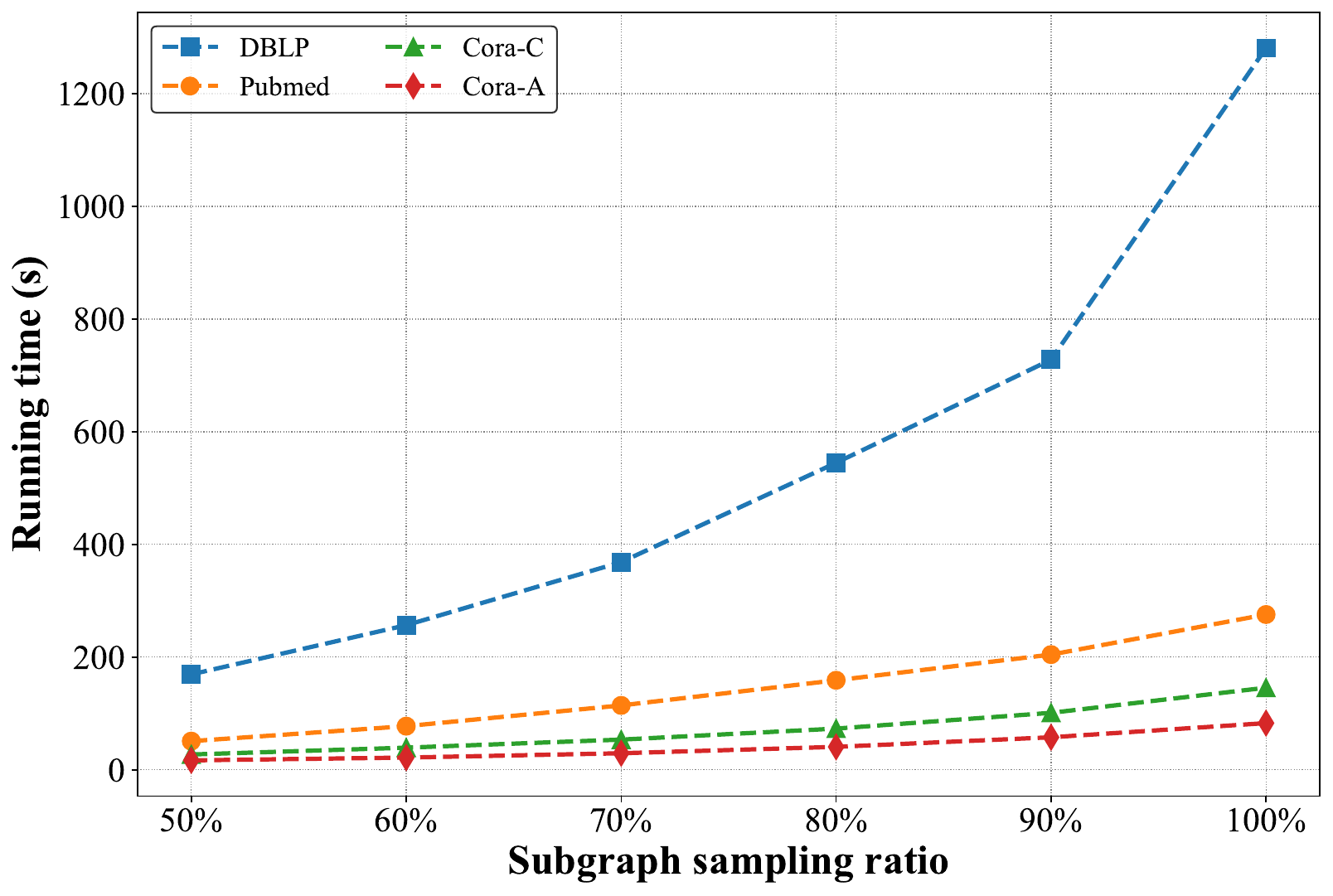}
    \caption{Scalability  of CAHC on the DBLP, cora-C, Pubmed, and Cora-A datasets.
    }
    \label{fig:scalability_analysis}
\end{figure}

To evaluate the scalability of CAHC, we randomly sample hypergraphs of different sizes from the DBLP, Cora-C, Pubmed, and Cora-A datasets and test CAHC on these extracted hypergraphs. For each dataset and each sampling ratio in {50\%, 60\%, 70\%, 80\%, 90\%, 100\%}, we extract hypergraphs containing the corresponding ratio of nodes from the dataset, with hyperedges that include the extracted nodes. The runtime of CAHC on these datasets is shown in Figure \ref{fig:scalability_analysis}.

\begin{figure}[htbp!]
  \centering
  \begin{subfigure}[b]{0.95\columnwidth}
    \centering
    \includegraphics[width=\textwidth]{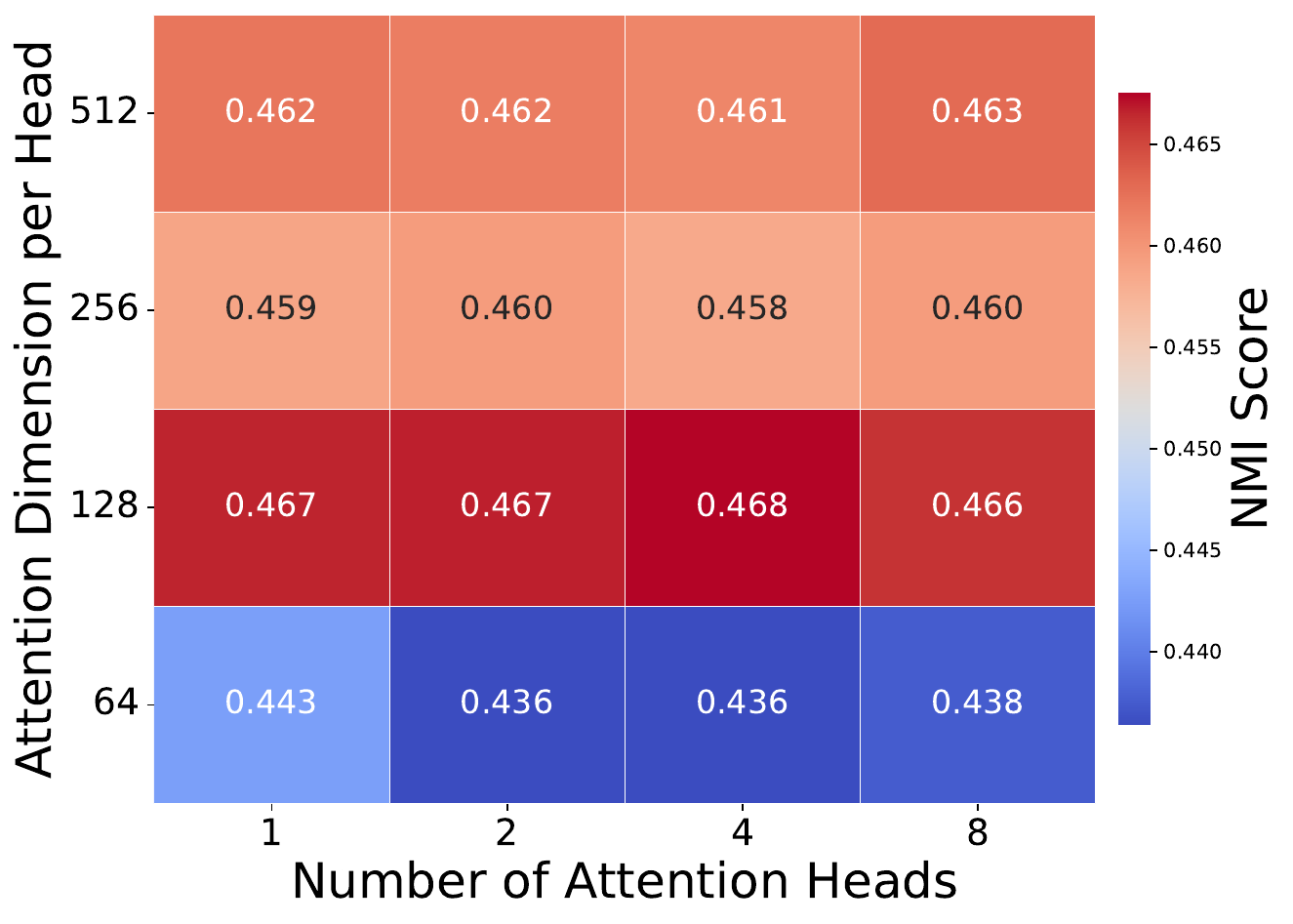}
    \caption{Citeseer}
    \label{fig:heatmap_citeseer}
  \end{subfigure}%
  
  \begin{subfigure}[b]{0.95\columnwidth}
    \centering
    \includegraphics[width=\textwidth]{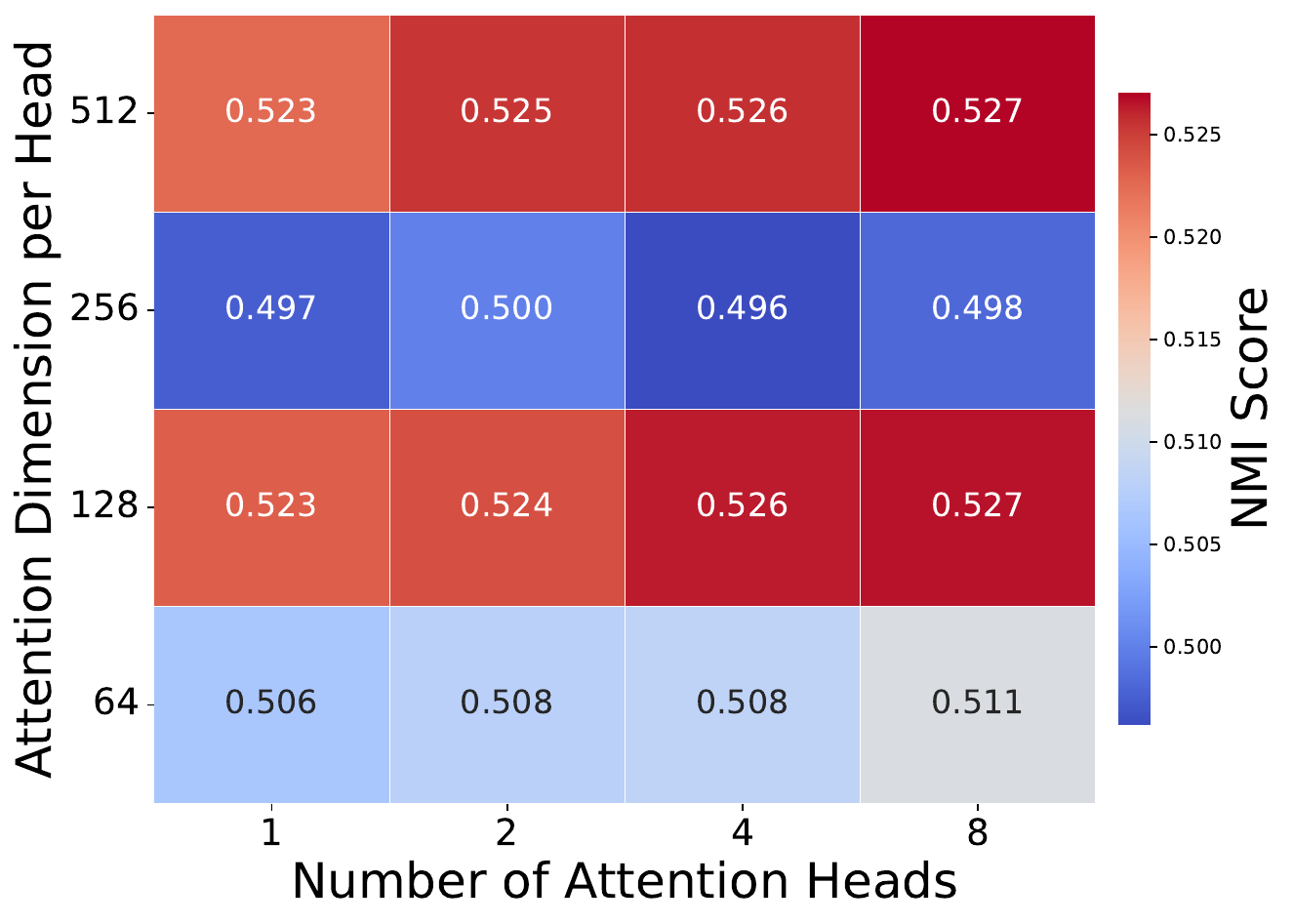}
    \caption{Cora-A}
    \label{fig:heatmap_cora_coauthor}
  \end{subfigure}
  \caption{Parameter analysis of the number of heads and attention dimension on the Citeseer and Cora-A datasets. The color intensity corresponds to the NMI score.}
  \label{fig:param_analysis_heatmaps}
\end{figure}

\begin{figure*}[!t]
  \centering
  \begin{subfigure}[b]{0.32\textwidth}
    \centering
    \includegraphics[width=\textwidth]{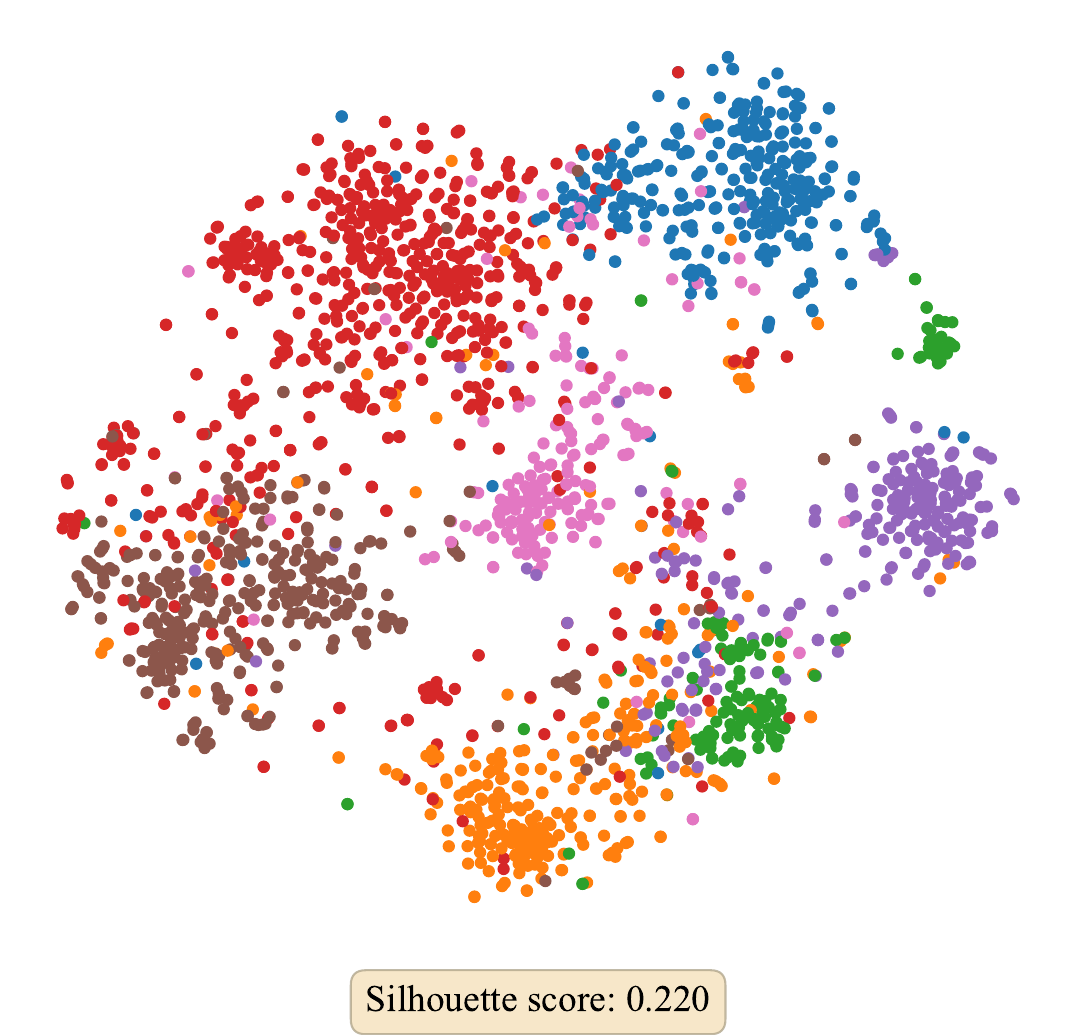}
    \caption{CAHC (Cora-A)}
    \label{fig:viz_cora_coauthor_ours}
  \end{subfigure}
  \hfill
  \begin{subfigure}[b]{0.32\textwidth}
    \centering
    \includegraphics[width=\textwidth]{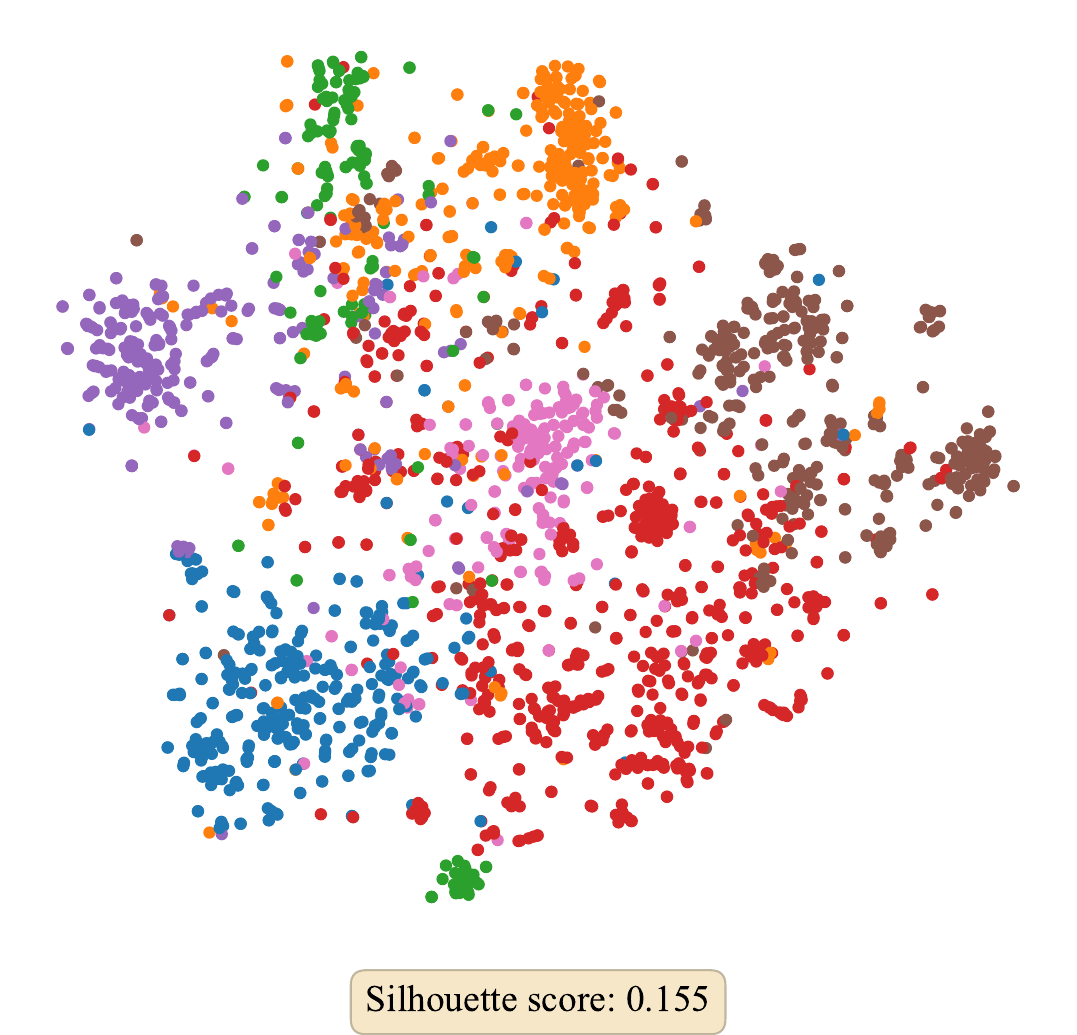}
    \caption{TriCL (Cora-A)}
    \label{fig:viz_cora_coauthor_tricl}
  \end{subfigure}
  \hfill
  \begin{subfigure}[b]{0.32\textwidth}
    \centering
    \includegraphics[width=\textwidth]{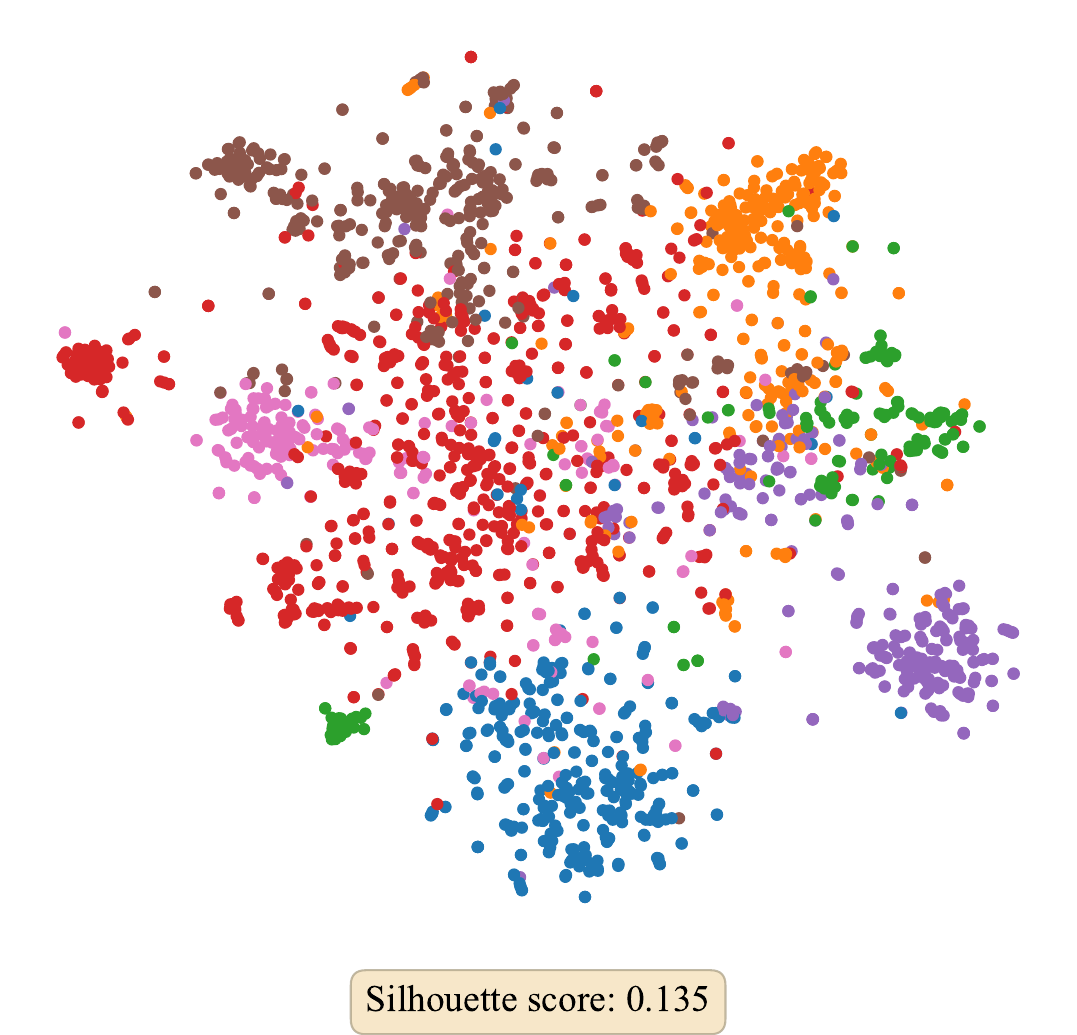}
    \caption{SE-HSSL (Cora-A)}
    \label{fig:viz_cora_coauthor_srhssl}
  \end{subfigure}

  \begin{subfigure}[b]{0.32\textwidth}
    \centering
    \includegraphics[width=\textwidth]{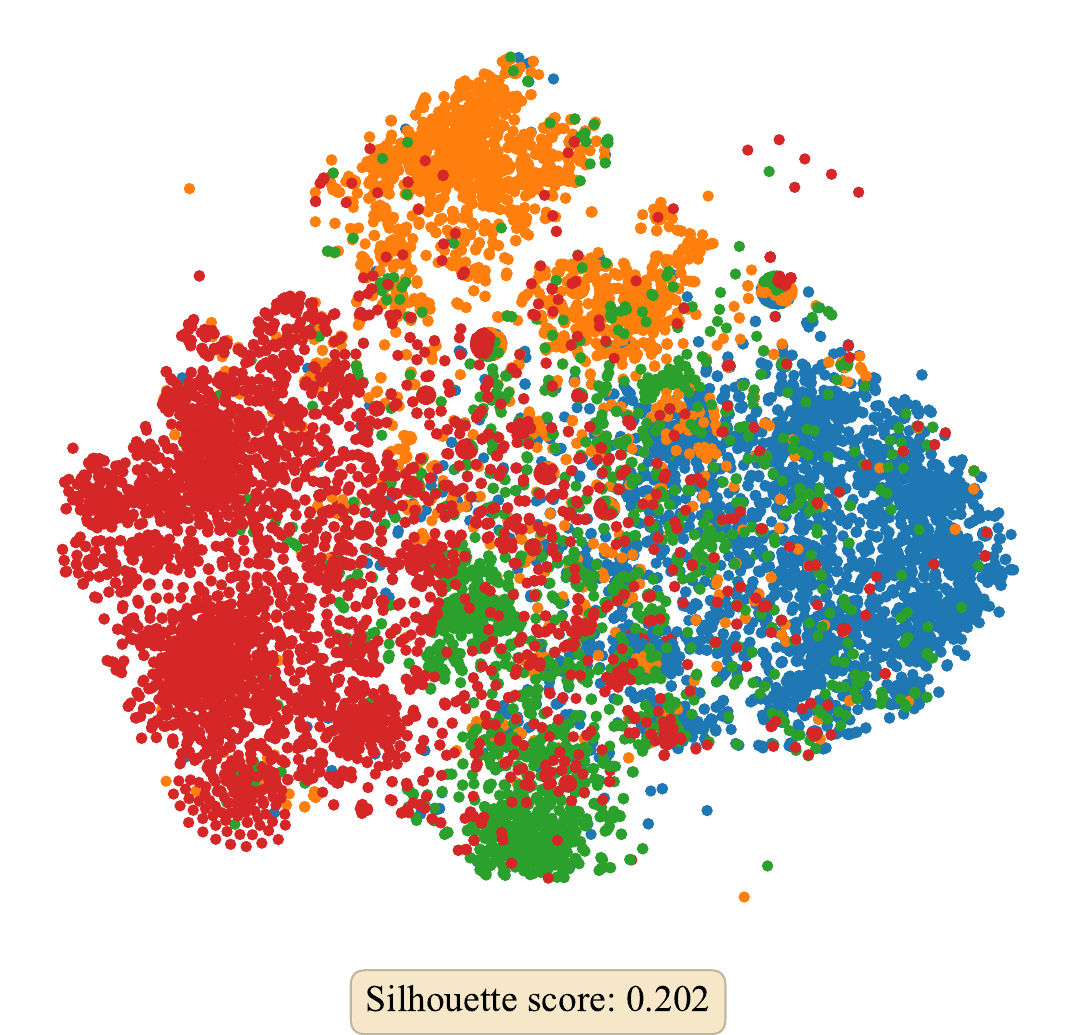} 
    \caption{CAHC (20NewsW100)}
    \label{fig:viz_20newsW100_ours} 
  \end{subfigure}
  \hfill
  \begin{subfigure}[b]{0.32\textwidth}
    \centering
    \includegraphics[width=\textwidth]{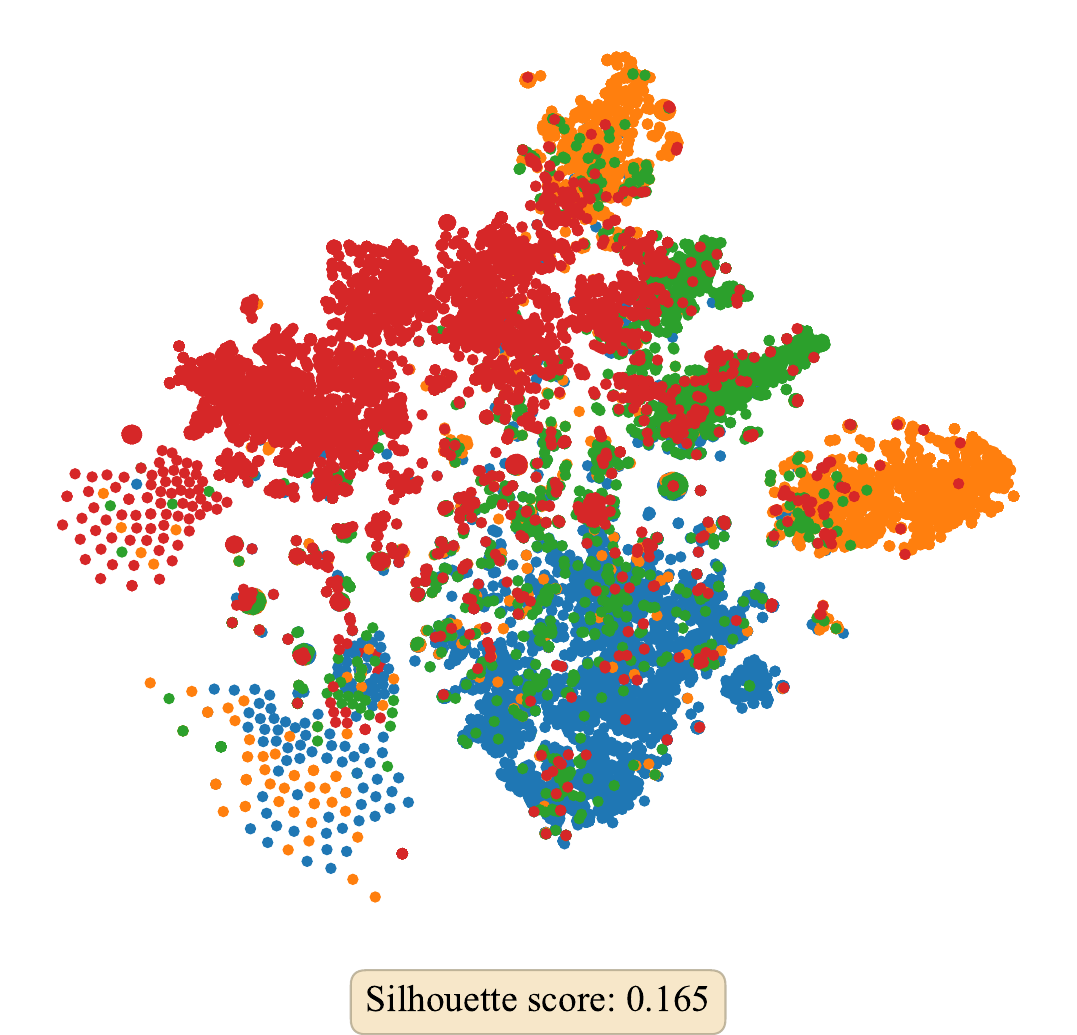}
    \caption{TriCL (20NewsW100)}
    \label{fig:viz_20newsW100_tricl} 
  \end{subfigure}
  \hfill
  \begin{subfigure}[b]{0.32\textwidth}
    \centering
    \includegraphics[width=\textwidth]{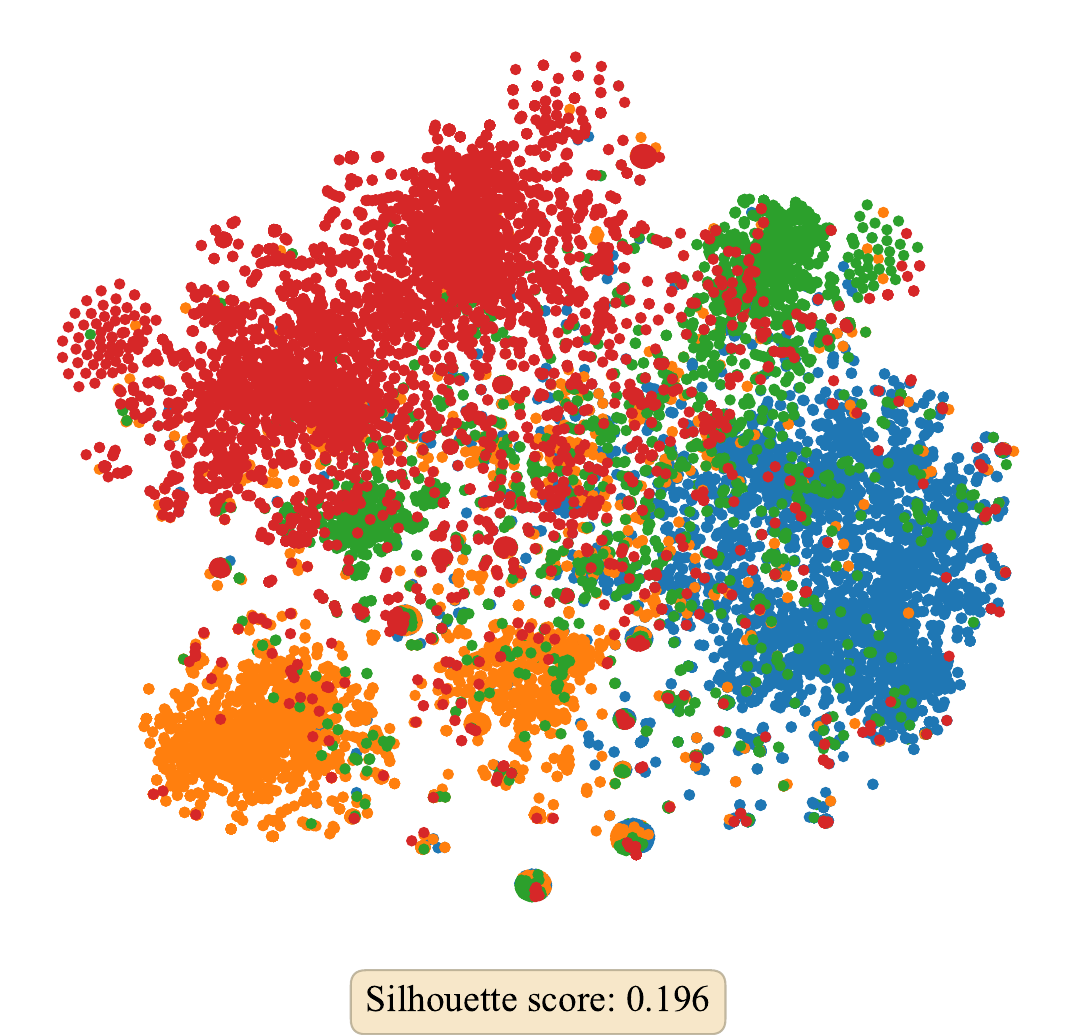}
    \caption{SE-HSSL (20NewsW100)}
    \label{fig:viz_20newsW100_srhssl}
  \end{subfigure} 
  \caption{
    t-SNE visualization of learned node embeddings on the \textbf{Cora-A} (top row) and \textbf{20newW100} (bottom row) datasets.
    The visualization compares the clustering quality of our proposed model against baselines. 
    Points represent nodes, colored by their ground-truth labels. 
  }
  \label{fig:viz_comparison_cora_coauthor} 
\end{figure*}

 Figure \ref{fig:scalability_analysis} shows that, as the percentage of nodes increases, the runtime of CAHC improves on datasets.
 This is because the computational complexity is directly related to the number of nodes, hyperedges, and degrees in the hypergraph. As the hypergraph grows in size, the computational cost of each message passing increases. For instance, as the number of nodes and hyperedges in the network increases, the average node degree rises, causing the hypergraph neural network to aggregate information from more neighbors, thus increasing computational cost.
\section{Additional experiments}
\subsection{Backbone encoder}
\label{sec:appendix_encoder_sensitivity}

To evaluate the number of attention heads and the dimension of each head on the performance of CAHC, 
we conduct a parameter analysis on the  Citeseer and Cora-Ar datasets.
The number of heads was varied within the set $\{1, 2, 4, 8\}$, while the attention dimension was selected from $\{64, 128, 256, 512\}$. The performance, measured by Normalized Mutual Information (NMI), is visualized via heatmaps for the Citeseer and Cora-A datasets in Figure~\ref{fig:param_analysis_heatmaps}.

Figure~\ref{fig:param_analysis_heatmaps} shows that CAHC on both datasets achieves the best performance when the dimensionality is moderate (128), and the model performance seems to be more sensitive to the choice of attention dimension than the choice of the number of heads. Furthermore, we observe that a multi-head configuration generally outperforms a single-head setting. This validates the effectiveness of the multi-head mechanism, suggesting that allowing the model to attend to information from different representation subspaces simultaneously may be beneficial for capturing diverse features and ultimately enhancing clustering performance.

\subsection{Qualitative analysis}
To intuitively compare the quality of the learned embeddings of algorithms, we employed t-SNE \cite{MaatenVisualizing2008} for visualizing the node representations learned by CAHC, TriCL,  and SE-HSSL. 
Figure~\ref{fig:viz_comparison_cora_coauthor} presents the t-SNE visualizations of node embeddings on the Cora-A and 20NewsW100 datasets.

Figure~\ref{fig:viz_comparison_cora_coauthor} shows that while TriCL and SE-HSSL yield disorganized results with no discernible clusters, CAHC demonstrates compact and well-separated clusters. This visual assessment is confirmed quantitatively by the Silhouette score \cite{RousseeuwSilhouettes1987}.
One potential reason TriCL and SE-HSSL  fail to produce separable embeddings is that they lack clustering guidance.
On the Cora-A dataset, CAHC achieves the highest Silhouette score, indicating the best cluster quality. 
These conclusions are consistent with results in Table~\ref{tab:main_results_acm_style}.

\end{document}